\documentclass[letterpaper]{article} 
\PassOptionsToPackage{table,xcdraw}{xcolor}
\usepackage{aaai2026}  
\usepackage{times}  
\usepackage{helvet}  
\usepackage{courier}  
\usepackage[hyphens]{url}  
\usepackage{graphicx} 
\usepackage{natbib}  
\usepackage{caption} 
\usepackage[ruled,vlined,linesnumbered]{algorithm2e}
\usepackage{newfloat}
\usepackage{listings}
\usepackage{soul} 
\usepackage{times}
\usepackage{latexsym}
\usepackage[T1]{fontenc}
\usepackage[utf8]{inputenc}
\usepackage{microtype}
\usepackage{inconsolata}
\usepackage{graphicx}
\usepackage{amsmath}
\usepackage{graphicx}
\usepackage{amssymb}
\usepackage{paralist}
\usepackage{booktabs}
\usepackage{textcomp}

\usepackage{multirow}
\usepackage{makecell}
\usepackage{subfigure}
\usepackage{float}
\definecolor{lightgray}{gray}{0.93}
\usepackage{amsthm}
\newtheoremstyle{myboldhead}
  {3pt}                 
  {3pt}                 
  {\normalfont}         
  {}                    
  {\bfseries}           
  {.}                   
  { }                   
  {}                    

\theoremstyle{myboldhead}
\newtheorem{definition}{Definition}
\newtheorem{assumption}{Assumption}

\newcommand{\red}[1]{\textcolor{black}{{#1}}}

\DeclareCaptionStyle{ruled}{labelfont=normalfont,labelsep=colon,strut=off} 
\floatstyle{ruled}
\newfloat{listing}{tb}{lst}{}
\floatname{listing}{Listing}
\title{Accommodate Knowledge Conflicts in Retrieval-augmented LLMs: Towards Robust Response Generation in the Wild}
\author{
    Jiatai Wang\textsuperscript{\rm 1,\rm 2}, 
    Zhiwei Xu\textsuperscript{\rm 2,\rm 3}\thanks{Corresponding authors}, 
    Di Jin\textsuperscript{\rm 5}, 
    Xuewen Yang\textsuperscript{\rm 4}, 
    Tao Li\textsuperscript{\rm 1,\rm 2}\footnotemark[1]
}
\affiliations{
    \textsuperscript{\rm 1}The College of Computer Science, Nankai University, Tianjin, China\\
    \textsuperscript{\rm 2}Haihe Lab of ITAI, Tianjin, China\\
    \textsuperscript{\rm 3}Institute of Computing Technology, Chinese Academy of Sciences, Beijing, China\\
    \textsuperscript{\rm 4}The Department of Electrical and Computer Engineering, Stony Brook University, New York, U.S.A\\
    \textsuperscript{\rm 5}Eigen AI, Palo Alto, U.S.A\\
    
    1120240357@mail.nankai.edu.cn, xuzhiwei2001@ict.ac.cn, di@eigenai.com, michelyang1990@gmail.com, litao@nankai.edu.cn
}

\usepackage{bibentry}

\begin{document}

\maketitle

\begin{abstract}
The proliferation of large language models (LLMs) has significantly advanced intelligent systems. Unfortunately, LLMs often face knowledge conflicts between internal memory and retrieved external information, arising from misinformation, biases, or outdated knowledge. These conflicts undermine response reliability and introduce uncertainty in decision-making. In this work, we analyze how LLMs navigate knowledge conflicts from an information-theoretic perspective and reveal that when conflicting and supplementary information exhibit significant differences, LLMs confidently resolve their preferences and alleviate the uncertainty during their response generation. When this difference is ambiguous, LLMs experience considerable uncertainty about their generation. Based on this insight, we propose Swin-VIB, a novel framework that integrates a pipeline of variational information bottleneck models to adapt the retrieved information difference, facilitating robust response generation of LLMs even in conflicting contexts. 
Extensive experiments confirm our theoretical analysis and demonstrate the performance of Swin-VIB. Notably, Swin-VIB outperforms all competitive baselines in terms of the accuracy of the multiple-choice task, while improving the EM values in the open-ended QA task by at least 11.14\%.
\end{abstract}


\section{Introduction}
\begin{figure}[t]
  \includegraphics[width=\columnwidth]{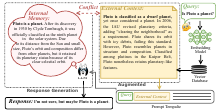}
  \caption{Illustration of knowledge conflict in RAG.}
  \label{intro-motivation}
\end{figure}

As large language models (LLMs) \cite{brown2020language, openai2023gpt} continue to proliferate, their powerful generation capabilities facilitate more transformative technologies for advanced intelligent systems.
Among these novel technologies, despite the response generation technique \cite{li2024matching} allows LLMs to answer queries without rigid retrieval, it still suffers from hallucinations.
Augmented LLMs with external knowledge, including retrieval augmented generation (RAG), have become a promising solution to mitigating these issues \cite{lewis2020retrieval, gao2023retrieval, asai2023self}. However, the disparity between the internal memory of LLM and the retrieved external context always leads to knowledge conflicts \cite{xu2024knowledge}.
These conflicts arise from misinformation, unreliable sources, and publisher bias in retrieved information, and the difficulty in synchronizing parametric knowledge with the context in real time further exacerbates these conflicts.
As illustrated in Figure \ref{intro-motivation}, these conflicts introduce uncertainty in response generation\cite{wu2024faithful}, posing a serious threat to the reliability of the responses and increasing the risk of biased or erroneous inference.

The existing approaches do not fully agree on how LLMs resolve knowledge conflicts. Some works prefer to fine-tune or edit LLMs according to external context and guiding response generation. \cite{shi2023context,jin2024cutting,zhang2024truthx}. In contrast, others attach external-validation modules that merely admit trusted context to supplement the internal memory of LLMs \cite{yu2024truth,kortukov2024studying}.
However, both of them have a fixed preference for the used knowledge that is often violated. 
This contradiction exposes a fundamental dilemma that without a principled framework explaining how such preferences emerge, either fine-tuning or external-validation modules risks the wrong optimizing results.
Although new efforts try to re-balance internal memory and external context during model decoding \cite{shi2023trusting,jin2024tug,huang2025pip}, their token-level weight updating fails to scale to mitigate fundamental knowledge conflicts in practice. 
Therefore, a theoretical analysis of the knowledge preference and uncertainty is highly desired to extend the existing heuristic guidelines toward knowledge conflicts mitigation and robust response generation.

To address this challenge, we re-examine knowledge conflict in response generation from the perspective of information theory. 
We analyze how LLMs establish preferences on the information used to generate responses when faced with conflicting knowledge sources and observe:
\begin{compactitem}
    \item When the disparity between conflicting and supplementary information is significant, LLMs confidently settle into a rational knowledge preference and generate reliable responses.
    \item  When the distinction is ambiguous, LLMs experience extreme uncertainty, making their response unreliable.
\end{compactitem}

Inspired by these insights, we propose Swin-VIB, a Variational Information Bottleneck (VIB) approach with a Sliding window. More specifically, we leverage a pipeline-based multiple variational information bottleneck models \cite{alemi2016deep} to augment the retrieved information adaptively and guide the preference of LLMs. This design adapts the LLM’s preference and minimizes ambiguity, enabling more accurate and consistent response generation even in challenging scenarios with knowledge conflicts. Swin‑VIB can be plugged into RAG pipelines almost without any additional overhead.
Our major contributions are summarized as follows:

\begin{compactitem}
    \item We model the interplay between LLMs' internal memory and external context and release the preference principle behind knowledge conflict in retrieval-augmented LLMs. With significant differences between conflicting and supplementary information, LLMs have more confidence to settle into a rational preference; otherwise, LLMs fall into extremely high uncertainty.

    \item This analysis of LLMs' preference provides insight into accommodating knowledge conflicts in retrieval-augmented LLMs. 
    In this way, we propose Swin-VIB\footnote{The code is available at https://github.com/JiataiWang/Swin-VIB.}, a sliding-window approach that integrates multiple variational information bottlenecks to adapt perplexing knowledge, enhancing it to guide LLMs toward accurate responses.

    \item Extensive experiments across multiple-choice and open-ended question answering (QA) validate our theoretical result and demonstrate that Swin-VIB outperforms baselines for robust response generation in the wild.
\end{compactitem}
To the best of our knowledge, this is the first approach that analyzes and evaluates knowledge conflicts of retrieval-augmented LLMs. A robust response generation paradigm is proposed according to the obtained insight, rather than merely relying on empirical evaluations, thus enjoying enhanced performance.
\section{Preliminary}

\subsection{Definitions}
Knowledge conflicts may cause uncertainty in response generation, where LLMs need to make a choice with conflicting knowledge from their internal memory and the external context. Considering the black-box nature of LLMs, it remains challenging to analyze the fundamental mechanism of these conflicts between invisible internal knowledge and external contexts.
Without a theoretical analysis of knowledge conflicts, the limitations of empirical rules and experimental settings in response generation can hardly alleviate knowledge conflicts \cite {wu2024faithful, xie2023adaptive, jin2024tug}. 
To tackle this challenge, we propose a theoretical framework that shows knowledge conflicts can be defined in terms of conditional entropy. The corresponding symbol system is outlined in Table \ref{tab:notion}. 
\begin{table}[h]
\renewcommand\arraystretch{0.5}
\centering
\setlength{\tabcolsep}{0.5mm}{
\begin{tabular}{ll}
\hline
Symbol & Meaning \\ \hline
$Q$ & Queries \\ \hline
$R=B(Q)$      & \begin{tabular}[c]{@{}l@{}}External contexts from the \\ knowledge base $B$\end{tabular} \\ \hline
$O = \text{LLM}(R,Q)$ & Responses generated by the LLM \\ \hline
\end{tabular}}
\caption{A list of Symbols}
\label{tab:notion}
\end{table}
\begin{definition}[\textit{\textbf{Uncertainty of response generation}}]\hfill\\
\label{eq:definition:Uncertainty}
\textit{The uncertainty of $O$ given the $Q$ and $R$ is represented by conditional entropy $\mathbb{H}$, is formalized by}
\begin{equation}
\label{definition:Uncertainty}
\mathbb{H}(O \mid R, Q)=-\sum_{o\in O, r \in R, q \in Q} p(o, r, q) \log p(o \mid r, q),
\end{equation}
\end{definition}
A commonsense assumption is taken to regulate the RAG process.
\begin{assumption}[\textit{\textbf{Non-void Retrieval}}]
\label{assumption-none}
\textit{A qualified external information retriever can recall external contexts correlated with the corresponding query~\cite{robertson1976relevance}, that is $\forall q \in Q$, $\exists r\in R$, $p(r, q) > 0$, where $q$ is a query, and $r$ is one correlated external context.} 
\end{assumption}

\subsection{Theoretical Analysis}\label{sec:Theoretical_Analysis}
We reveal how the difference between conflicting and supplementary information influences the reliability of LLMs. A brief analysis is listed as follows, and all details about this analysis are provided in Appendix~A.\\
\textit{\textbf{Step 1: Uncertainty formulation}}. According to Definition \ref{definition:Uncertainty} and Assumption \ref{assumption-none}, we rewrite Formula \ref{eq:definition:Uncertainty} based on the law of total probability and the chain rule (see Appendix~A.1): 
\begin{equation}
\label{eq:entropy-final-short}
\begin{aligned}
\mathbb{H}(O\mid R,Q)
=\sum_{q} p(q)\sum_{r} p(r\mid q)\sum_{o}\psi\!\bigl(p(o\mid r,q)\bigr),
\end{aligned}
\end{equation}
where $\psi(\cdot)$ denotes instance-level uncertainty, and is calculated by
\begin{equation}
\label{eq-2}
\psi(p(o\mid r, q))=-p(o\mid r, q)\log p(o\mid r, q)
\end{equation}\\
\textit{\textbf{Step 2: Latent-state decomposition.}}
Let $X$ be the high-dimensional latent space of the LLM, and let $x\!\in\!\mathbf X$ denotes a specific latent state activated during generation (see Appendix A.2). We have:
\begin{equation}
\label{fm:p()}
\begin{aligned}
&p(o \mid r, q)= \int_{X} p(o \mid r, q, x) p(x\mid r,q)\,dx\\ 
&= \int_{X} p(o \mid r, q, x) \frac{p(r,q\mid x)p(x)}{p(r, q)}\,dx\\
&\propto \int_{X} p(o \mid r, q, x) p(r,q\mid x)p(x)\,dx\\
&\propto \int_{X} p(o \mid r, q, x) \frac{p(r,q\mid x)}{p(r,q\mid x_{\gamma})} p(x)\,dx\\
&\approx\int_{X} p\left(o \mid r, q, x\right) exp[\log{p\left(r \mid x\right)}- \\
&\qquad \qquad \qquad \qquad \log{p\left(r \mid x_{\gamma}\right)}]p(x)dx\\
&\propto \int_{X} p(o\mid r,q,x)\,\exp\!\bigl[I_c-I_s\bigr]\,p(x)\,dx,
\end{aligned}
\end{equation}
where self-information $I_s=-\log p(r\mid x)$ denotes the external information consistent with the LLM’s memory, namely, supplementary information, whereas $I_c=-\log p(r\mid x_\gamma)$ denotes the information that contradicts internal memory and is defined as conflicting information, with $x_\gamma\in X$ being a latent state aligned to the retrieved context~$r$.\\
\textit{\textbf{Step 3: Approximation}}  
Leveraging the monotonicity of $\exp(\cdot)$, and the constants are reduced after normalizing $p(o\mid r,q)$ over $o$, we apply a Taylor expansion and arrive at the following approximation (see Appendix~A.3):
\begin{equation}
\label{eq:linear-proxy}
p(o\mid r,q)\;\propto\;I_c - I_s .
\end{equation}
Equation~\eqref{eq:linear-proxy} is a first-order proxy of the exponential, and monotonic claims in Step~4 rely only on the monotonicity of the exponential tilt (proved in Appendix~A.3).\\
\textit{\textbf{Step 4: Theoretical statement.}}
\begin{figure}[t]
\centering
  \includegraphics[width=0.9\linewidth]{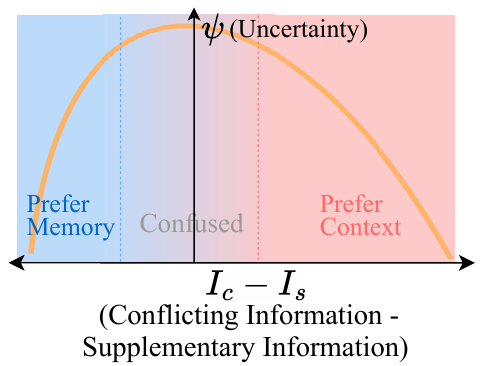}
  \caption{Relationship between uncertainty and the information difference.}
  \label{fig:detecting: proportion}
\end{figure}
According to Formula \eqref{eq:linear-proxy}, Formula \eqref{eq-2} reveals a relation between uncertainty $\psi$ and the information difference $\Delta I = I_c - I_s$.  
Larger $|\Delta I|$ consistently corresponds to lower $\psi$, \textit{i.e.} $|\Delta I|\!\uparrow \Longleftrightarrow \psi \!\downarrow$, which can also be found in Figure~\ref{fig:detecting: proportion}. 
Equivalently, the conditional entropy is the weighted expectation of instance-level
uncertainty,
\begin{equation}
\label{eq}
\mathbb{H}(O\!\mid\!R,Q)=\mathbb{E}_{q\sim p(q)}\,\mathbb{E}_{r\sim p(r\mid q)}
\Bigl[\sum_o \psi\bigl(p(o\mid r,q)\bigr)\Bigr],
\end{equation}
So lowering the per $(q,r)$ uncertainty in expectation lowers $\mathbb{H}(O\!\mid\!R,Q)$.
The full derivation can be found in Appendix~A.4.

\begin{figure*}[t]
  \centering
  \includegraphics[width=1\linewidth]{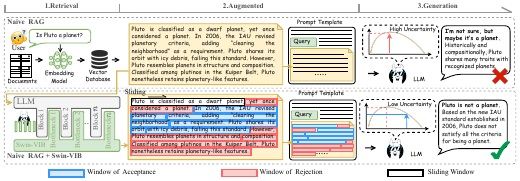}
  \caption{An overview of response generation with Swin-VIB.}
  \label{fig:mitigating:inference}
\end{figure*}

\section{Methodology}
 According to our theoretical analysis, a larger information difference $\lvert\Delta I\rvert$ of conflicting and supplementary information enhances response reliability, and thus we take this insight to design Swin-VIB, which integrates a pipeline of variational information bottleneck models to adapt the external information in those context windows with large $\lvert\Delta I\lvert$.

\subsection{Overview}
To maximize the information difference between conflicting and supplementary information in the external context, the retrieved context is first segmented into fixed-length windows to give a unified unit on which the information difference is measured. Specifically, Swin-VIB slides over these windows and quickly predicts the difference in information contained in each window. The window with a large enough information difference is accepted; otherwise, the window is rejected. Finally, Swin-VIB concatenates the accepted windows with the query to form a prompt, as shown in Figure \ref{fig:mitigating:inference}.

\subsection{Data Preparation}
\label{sec:data}
Given a dataset $\mathcal D=\{(q,r)\}$, where each query $q$ is paired with an external context $r\in\{r_c,r_s,r_m\}$, $r_c$ is conflicting context, $r_s$ is supplementary context, and $r_m$ is their mixture. To construct the mixed context $r_m$, conflicting and supplementary contexts interleave each other by every 4 tokens, eliminating the possibility of sampling a window only with conflicting context or supplementary context.  
For each query $q$, we randomly extract a fixed-length window of length 7 from the context $r$.
\begin{equation}
\omega^* \leftarrow \text{RandomWindow}(r,\text{len}=7), 
\end{equation}
and for this window, a binary label is assigned as
\begin{equation}
\label{fm:y}
\mathbf{Y} = 
\begin{cases} 
1, & \text{if } r \text{ is from single source, }r\in\{r_c,r_s\} \\
\begin{array}{@{}c@{}}
0 
\end{array}, & 
\begin{array}{@{}l@{}}
\text{if } r \text{ has multiple sources}, r=r_m 
\end{array}
\end{cases}
\end{equation}
This setting guarantees $\mathbf{Y}$ represents the information diffidence $\lvert\Delta I\lvert$. 
Because $r \in \{ r_c, r_s\}$ denotes all tokens originate from a single source, the $\omega^*$ carries a maximal difference between conflicting and supplementary information. If $r \in \{ r_m\}$, the $\omega^*$ inevitably contains multiple optional instances from different sources and therefore exhibits a small information difference, matching the multiple optional condition.

Feeding each $\omega^{\!*}$ into a frozen LLM yields a stack of per-layer attention matrices $\{\mathbf A_n\}$.  
We average heads to obtain  
\begin{equation}
\mathbf G_n \;=\; \textsc{MeanHeads}(\mathbf A_n),
\end{equation}
and construct the training set
\begin{equation}
T_n \;=\; \bigl\{\mathbf G_n,\,\mathbf Y\bigr\}^{N}
\end{equation}
All $\mathbf{G}_n$ of LLM decoder layers share the same $\mathbf Y$. Detailed in Algorithm~\ref{alg:train-construct}.
\begin{algorithm}[t]
\caption{Training-Set Construction}
\label{alg:train-construct}
\KwIn{Dataset $\mathcal{D}$; LLM $\mathcal{M}$ with $N$ layers\\}
\KwOut{Training sets $T_n=(\mathbf{G},\mathbf{Y})_{n=1}^{N}$}
\BlankLine
Initialise $T_n\leftarrow[\,]$ for $n=1,\dots,N$\;
\For{each sample $\{q, r\}
\in\mathcal{D}$}{
    $\omega^{*} \leftarrow\text{RandomWindow}(r,\,\text{len})$\;
    $\mathbf Y \leftarrow 
       \begin{cases}
          1,& \text{if } r\in\{r_c,r_s\}\\
          0,& \text{if } r=r_m
       \end{cases}$\;
    \For{$n\leftarrow1$ \KwTo $N$}{
        $\mathbf A^{(n)}\leftarrow\text{Attention}(\omega^{*},n)$\;
        $\mathbf G_{n}\leftarrow\text{MeanHeads}(\mathbf A_{n})$\;
        \textbf{append} $(\mathbf G_{n},Y)$ to $T_n$\;
    }
}
\Return{$\{T_n\}_{n=1}^{N}$}\tcp*[r]{each saved via \texttt{pickle.dump}}
\end{algorithm}
\begin{figure}[t]
\centering
  \includegraphics[width=1\columnwidth]{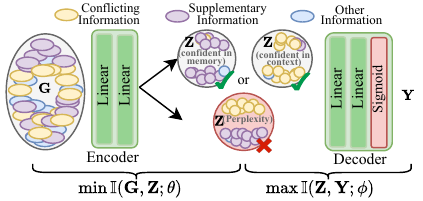}
  \caption{Bottleneck model structure and training objective.}
  \label{fig:mitigating:training}
\end{figure}

\subsection{Swin-VIB Architecture \& Training}
\label{sec:arch}
We propose a novel method Swin-VIB to compress various information and extract their differences adaptively. Inspired by the information bottleneck structure \cite{tishby2000information}, Swin-VIB introduces an information bottleneck into each transformer decoder layer to predict the information difference of the external context. In the following, we describe how the constructed training sets $T_n$ are fed into these bottlenecks for training.

As shown in Figure \ref{fig:mitigating:training}, a bottleneck model consists of an encoder and a decoder, the encoder projects $\mathbf G_n$ of an LLM layer to the mean $\boldsymbol\mu_n$ and log-variance $\log\boldsymbol\sigma_n^2$ of each Gaussian latent feature. 
\begin{equation}
q_{\theta}(\mathbf Z_n\mid\mathbf G_n)=\mathcal N\!\bigl(\boldsymbol\mu_n,\operatorname{diag}(\boldsymbol\sigma_n^{2})\bigr).
\end{equation}
where $\theta$ and $\phi$ are the parameters of the encoder and the decoder. Via the reparameterization trick \cite{kingma2013auto}, we obtain a differentiable latent representation, 
\begin{equation}
\mathbf Z_n=\boldsymbol\mu_n+\boldsymbol\sigma_n\odot\boldsymbol\epsilon,\qquad
\boldsymbol\epsilon\sim\mathcal N(\mathbf 0,\mathbf I).
\end{equation}
The decoder maps $\mathbf Z_n$ to the logit of $\hat Y_n=p_{\phi}(Y=1\mid\mathbf Z_n)$, and predicts whether a window includes a larger information difference.
Each bottleneck is trained independently by minimizing the information bottleneck loss:
\begin{equation}
\label{eq:ib-loss}
\begin{aligned}
\mathcal L_n(\theta,\phi)=
&\underbrace{\mathbb E_{q_{\theta}}\bigl[-\log p_{\phi}(Y\mid\mathbf Z_n)\bigr]}_{\uparrow \mathbb I(\mathbf Z_n;Y)}
+\\
&\underbrace{\beta\mathrm{KL}\bigl(q_{\theta}(\mathbf Z_n\mid\mathbf G_n)|p(\mathbf Z)\bigr)}_{\downarrow \mathbb I(\mathbf G_n;\mathbf Z_n)}
\end{aligned}
\end{equation}
where $p(\mathbf Z)=\mathcal N(\mathbf 0,\mathbf I)$ is an isotropic prior and parameter $\beta$ controls the compression–prediction trade-off.  
It is by minimizing the mutual information $\mathbb I(\mathbf{G}, \mathbf{Z})$ to guide the encoder to adaptively learn the key features to identify information differences. The KL term upper-bounds $\mathbb I(\mathbf G_n;\mathbf Z_n)$, and thus makes the encoder discard redundant information.

\subsection{Robust Response Generation}
\label{sec:inference}
To achieve robust response generation, we incorporate the corresponding bottleneck into the inference process of each transformer-decoder layer of LLM. More specifically, the RAG retriever  recalls an external context $r$, which is partitioned into fixed-length windows $r=\{\omega^{1},\dots,\omega^{K}\}$. Each window $\omega^{k}$ is fed to the frozen LLM to get attention representation $\mathbf G_{n}^{k}$, which is collected from every decoder layer. The representation is fed into the corresponding layer-specific bottleneck model, where the encoder’s mean $\boldsymbol\mu$ is efficiently taken as the latent $\mathbf Z_{n}$ to skip sampling. The bottleneck returns a layer-wise probability $p_{\phi_{n}}(Y{=}1\mid\mathbf G_{n}^{k})$ that predicts the information difference in terms of different context windows. The outputs of all bottlenecks are weighted to achieve an averaging result to guide robust response generation:
\begin{equation}
\hat p(q,\omega^{k})=\frac1N\sum_{n=1}^{N}p_{\phi_{n}}\!\bigl(Y{=}1\mid\mathbf G_{n}^{k}\bigr),
\end{equation}
A window is accepted if $\hat p(\omega^{k})\ge\xi$, where $\xi$ is a threshold. All accepted windows are concatenated with the query to construct the final prompt $[q;\mathcal \{\omega^{k}\mid\hat p(\omega^{k})\ge\xi\}]$. In this way, the LLM is guided to generate reliable responses, even in the case that the context contains a large amount of conflicting information.

\section{Experiments}
\label{experiments}
\begin{table*}[t]
\renewcommand\arraystretch{0.8}
\centering
\scriptsize
\setlength{\tabcolsep}{2mm}{
\begin{tabular}{cccccccccccccc}
\Xhline{1.2pt} 
\multirow{2}{*}{LLM}       & \multirow{2}{*}{Method} & \multicolumn{6}{c}{ConflictQA} & \multicolumn{6}{c}{DRUID} \\ \cline{3-14} 
                           &                                    & ACC$\uparrow$                        & CR$\uparrow$                         & RR$\uparrow$                         & Mean-$\psi\downarrow$               & UAR$\downarrow$                      & TRE$\downarrow$                       & ACC$\uparrow$                       & CR$\uparrow$                         & RR$\uparrow$                         & Mean-$\psi\downarrow$                & UAR$\downarrow$                      & TRE$\downarrow$  \\ \hline
\multirow{6}{*}{Llama2-7B} & \cellcolor[HTML]{F5F5F5}Closed-book&\cellcolor[HTML]{F5F5F5}20.61         &\cellcolor[HTML]{F5F5F5} -            &\cellcolor[HTML]{F5F5F5} -            &\cellcolor[HTML]{F5F5F5} 0.31            & \cellcolor[HTML]{F5F5F5}-            &\cellcolor[HTML]{F5F5F5} -            &\cellcolor[HTML]{F5F5F5}50.58         &\cellcolor[HTML]{F5F5F5}-             &\cellcolor[HTML]{F5F5F5}-            &\cellcolor[HTML]{F5F5F5} 0.34              &\cellcolor[HTML]{F5F5F5} -            & \cellcolor[HTML]{F5F5F5}-     \\
                           & \cellcolor[HTML]{ECF4FF}In-context &\cellcolor[HTML]{ECF4FF}79.68         &\cellcolor[HTML]{ECF4FF}96.49         &\cellcolor[HTML]{ECF4FF}16.92         &\cellcolor[HTML]{FFF0F0}0.34          &\cellcolor[HTML]{FFF0F0}2.96          &\cellcolor[HTML]{FFF0F0}0.69          &\cellcolor[HTML]{ECF4FF}50.08         &\cellcolor[HTML]{ECF4FF}90.11         &\cellcolor[HTML]{ECF4FF}12.74        &\cellcolor[HTML]{FFF0F0}0.38         &\cellcolor[HTML]{FFF0F0}8.89          &\cellcolor[HTML]{FFF0F0}1.34   \\
                           & \cellcolor[HTML]{ECF4FF}CD$^{2}$   &\cellcolor[HTML]{ECF4FF}77.32         &\cellcolor[HTML]{ECF4FF}88.05         &\cellcolor[HTML]{ECF4FF}29.16         &\cellcolor[HTML]{FFF0F0}0.29          &\cellcolor[HTML]{FFF0F0}2.01          &\cellcolor[HTML]{FFF0F0}0.87          &\cellcolor[HTML]{ECF4FF}58.14         &\cellcolor[HTML]{ECF4FF}92.30         &\cellcolor[HTML]{ECF4FF}22.05       &\cellcolor[HTML]{FFF0F0}0.30          &\cellcolor[HTML]{FFF0F0}4.32          &\cellcolor[HTML]{FFF0F0}1.18    \\
                           & \cellcolor[HTML]{ECF4FF}Rowen-CL   &\cellcolor[HTML]{ECF4FF}80.10         &\cellcolor[HTML]{ECF4FF}92.59         &\cellcolor[HTML]{ECF4FF}28.55         &\cellcolor[HTML]{FFF0F0}0.28          &\cellcolor[HTML]{FFF0F0}1.30          &\cellcolor[HTML]{FFF0F0}0.79          &\cellcolor[HTML]{ECF4FF}52.35         &\cellcolor[HTML]{ECF4FF}94.76         &\cellcolor[HTML]{ECF4FF}17.30        &\cellcolor[HTML]{FFF0F0}0.29          &\cellcolor[HTML]{FFF0F0}5.92          &\cellcolor[HTML]{FFF0F0}1.26     \\
                           & \cellcolor[HTML]{ECF4FF}CK-PLUG    &\cellcolor[HTML]{ECF4FF}78.17         &\cellcolor[HTML]{ECF4FF}95.78         &\cellcolor[HTML]{ECF4FF}25.29         &\cellcolor[HTML]{FFF0F0}0.28          &\cellcolor[HTML]{FFF0F0}1.87          &\cellcolor[HTML]{FFF0F0}0.85         &\cellcolor[HTML]{ECF4FF}58.88         &\cellcolor[HTML]{ECF4FF}94.24         &\cellcolor[HTML]{ECF4FF}18.00       &\cellcolor[HTML]{FFF0F0}0.29          &\cellcolor[HTML]{FFF0F0}5.39         &\cellcolor[HTML]{FFF0F0}1.21     \\
                           & \cellcolor[HTML]{D5E3FB}Swin-VIB   &\cellcolor[HTML]{D5E3FB}\textbf{84.04}&\cellcolor[HTML]{D5E3FB}\textbf{98.18}&\cellcolor[HTML]{D5E3FB}\textbf{29.40}&\cellcolor[HTML]{F9CDCD}\textbf{0.24} &\cellcolor[HTML]{F9CDCD}\textbf{0.21} &\cellcolor[HTML]{F9CDCD}\textbf{0.64} &\cellcolor[HTML]{D5E3FB}\textbf{63.24}&\cellcolor[HTML]{D5E3FB}\textbf{94.87}&\cellcolor[HTML]{D5E3FB}\textbf{25.63}&\cellcolor[HTML]{F9CDCD}\textbf{0.26}&\cellcolor[HTML]{F9CDCD}\textbf{2.85} &\cellcolor[HTML]{F9CDCD}\textbf{1.09} \\ 
                           \hline
\multirow{6}{*}{Llama2-13B}& \cellcolor[HTML]{F5F5F5}Closed-book&\cellcolor[HTML]{F5F5F5}21.37         &\cellcolor[HTML]{F5F5F5} -            &\cellcolor[HTML]{F5F5F5} -           &\cellcolor[HTML]{F5F5F5} 0.30            & \cellcolor[HTML]{F5F5F5}-            &\cellcolor[HTML]{F5F5F5} -            &\cellcolor[HTML]{F5F5F5}50.93         &\cellcolor[HTML]{F5F5F5}-             &\cellcolor[HTML]{F5F5F5}-             &\cellcolor[HTML]{F5F5F5} 0.33            & \cellcolor[HTML]{F5F5F5}-     &\cellcolor[HTML]{F5F5F5} -\\
                           & \cellcolor[HTML]{ECF4FF}In-context &\cellcolor[HTML]{ECF4FF}79.83         &\cellcolor[HTML]{ECF4FF}97.33         &\cellcolor[HTML]{ECF4FF}16.02         &\cellcolor[HTML]{FFF0F0}0.33          &\cellcolor[HTML]{FFF0F0}1.98          &\cellcolor[HTML]{FFF0F0}0.82          &\cellcolor[HTML]{ECF4FF}51.73         &\cellcolor[HTML]{ECF4FF}92.21         &\cellcolor[HTML]{ECF4FF}11.19       &\cellcolor[HTML]{FFF0F0}0.37          &\cellcolor[HTML]{FFF0F0}7.16          &\cellcolor[HTML]{FFF0F0}1.29   \\
                           & \cellcolor[HTML]{ECF4FF}CD$^{2}$   &\cellcolor[HTML]{ECF4FF}80.20         &\cellcolor[HTML]{ECF4FF}93.52         &\cellcolor[HTML]{ECF4FF}27.23         &\cellcolor[HTML]{FFF0F0}0.28          &\cellcolor[HTML]{FFF0F0}4.85          &\cellcolor[HTML]{FFF0F0}0.88          &\cellcolor[HTML]{ECF4FF}60.72         &\cellcolor[HTML]{ECF4FF}93.12         &\cellcolor[HTML]{ECF4FF}18.55       &\cellcolor[HTML]{FFF0F0}0.29          &\cellcolor[HTML]{FFF0F0}4.04          &\cellcolor[HTML]{FFF0F0}1.15    \\
                           & \cellcolor[HTML]{ECF4FF}Rowen-CL   &\cellcolor[HTML]{ECF4FF}80.36         &\cellcolor[HTML]{ECF4FF}95.42         &\cellcolor[HTML]{ECF4FF}28.55         &\cellcolor[HTML]{FFF0F0}0.27          &\cellcolor[HTML]{FFF0F0}1.10          &\cellcolor[HTML]{FFF0F0}0.76          &\cellcolor[HTML]{ECF4FF}52.61         &\cellcolor[HTML]{ECF4FF}94.85         &\cellcolor[HTML]{ECF4FF}16.93       &\cellcolor[HTML]{FFF0F0}0.28          &\cellcolor[HTML]{FFF0F0}4.36          &\cellcolor[HTML]{FFF0F0}1.21     \\
                           & \cellcolor[HTML]{ECF4FF}CK-PLUG    &\cellcolor[HTML]{ECF4FF}81.44         &\cellcolor[HTML]{ECF4FF}97.27         &\cellcolor[HTML]{ECF4FF}27.38         &\cellcolor[HTML]{FFF0F0}0.27          &\cellcolor[HTML]{FFF0F0}1.78         &\cellcolor[HTML]{FFF0F0}0.78          &\cellcolor[HTML]{ECF4FF}61.07         &\cellcolor[HTML]{ECF4FF}\textbf{95.18}         &\cellcolor[HTML]{ECF4FF}18.83       &\cellcolor[HTML]{FFF0F0}0.28          &\cellcolor[HTML]{FFF0F0}5.47          &\cellcolor[HTML]{FFF0F0}1.19     \\
                           & \cellcolor[HTML]{D5E3FB}Swin-VIB   &\cellcolor[HTML]{D5E3FB}\textbf{85.68}&\cellcolor[HTML]{D5E3FB}\textbf{99.59}&\cellcolor[HTML]{D5E3FB}\textbf{30.40}&\cellcolor[HTML]{F9CDCD}\textbf{0.23} &\cellcolor[HTML]{F9CDCD}\textbf{0.18} &\cellcolor[HTML]{F9CDCD}\textbf{0.61} &\cellcolor[HTML]{D5E3FB}\textbf{63.43}&\cellcolor[HTML]{D5E3FB}95.16&\cellcolor[HTML]{D5E3FB}\textbf{25.48}&\cellcolor[HTML]{F9CDCD}\textbf{0.25}&\cellcolor[HTML]{F9CDCD}\textbf{1.96}&\cellcolor[HTML]{F9CDCD}\textbf{1.06} \\ 
                           \hline
\multirow{6}{*}{Llama2-70B}& \cellcolor[HTML]{F5F5F5}Closed-book&\cellcolor[HTML]{F5F5F5}20.99         &\cellcolor[HTML]{F5F5F5} -            &\cellcolor[HTML]{F5F5F5} -            &\cellcolor[HTML]{F5F5F5} 0.29            & \cellcolor[HTML]{F5F5F5}-            &\cellcolor[HTML]{F5F5F5} -            &\cellcolor[HTML]{F5F5F5}49.59        &\cellcolor[HTML]{F5F5F5}-             &\cellcolor[HTML]{F5F5F5}-             &\cellcolor[HTML]{F5F5F5} 0.32             & \cellcolor[HTML]{F5F5F5}-    &\cellcolor[HTML]{F5F5F5} - \\
                           & \cellcolor[HTML]{ECF4FF}In-context &\cellcolor[HTML]{ECF4FF}80.29         &\cellcolor[HTML]{ECF4FF}98.21         &\cellcolor[HTML]{ECF4FF}20.91         &\cellcolor[HTML]{FFF0F0}0.32          &\cellcolor[HTML]{FFF0F0}1.12          &\cellcolor[HTML]{FFF0F0}0.78          &\cellcolor[HTML]{ECF4FF}52.81         &\cellcolor[HTML]{ECF4FF}90.65         &\cellcolor[HTML]{ECF4FF}19.77       &\cellcolor[HTML]{FFF0F0}0.36           &\cellcolor[HTML]{FFF0F0}3.66          &\cellcolor[HTML]{FFF0F0}1.18   \\
                           & \cellcolor[HTML]{ECF4FF}CD$^{2}$   &\cellcolor[HTML]{ECF4FF}76.20         &\cellcolor[HTML]{ECF4FF}96.25         &\cellcolor[HTML]{ECF4FF}30.00         &\cellcolor[HTML]{FFF0F0}0.27          &\cellcolor[HTML]{FFF0F0}2.10          &\cellcolor[HTML]{FFF0F0}0.89          &\cellcolor[HTML]{ECF4FF}54.86         &\cellcolor[HTML]{ECF4FF}94.32         &\cellcolor[HTML]{ECF4FF}24.55       &\cellcolor[HTML]{FFF0F0}0.28          &\cellcolor[HTML]{FFF0F0}1.19          &\cellcolor[HTML]{FFF0F0}1.07    \\
                           & \cellcolor[HTML]{ECF4FF}Rowen-CL   &\cellcolor[HTML]{ECF4FF}81.43         &\cellcolor[HTML]{ECF4FF}97.04         &\cellcolor[HTML]{ECF4FF}29.46         &\cellcolor[HTML]{FFF0F0}0.26          &\cellcolor[HTML]{FFF0F0}0.95          &\cellcolor[HTML]{FFF0F0}0.74          &\cellcolor[HTML]{ECF4FF}54.44         &\cellcolor[HTML]{ECF4FF}95.60         &\cellcolor[HTML]{ECF4FF}23.89       &\cellcolor[HTML]{FFF0F0}0.26          &\cellcolor[HTML]{FFF0F0}0.97          &\cellcolor[HTML]{FFF0F0}1.06     \\
                           & \cellcolor[HTML]{ECF4FF}CK-PLUG    &\cellcolor[HTML]{ECF4FF}83.41         &\cellcolor[HTML]{ECF4FF}99.08         &\cellcolor[HTML]{ECF4FF}28.13         &\cellcolor[HTML]{FFF0F0}0.27         &\cellcolor[HTML]{FFF0F0}0.43          &\cellcolor[HTML]{FFF0F0}0.68         &\cellcolor[HTML]{ECF4FF}56.70         &\cellcolor[HTML]{ECF4FF}\textbf{97.16}         &\cellcolor[HTML]{ECF4FF}24.42       &\cellcolor[HTML]{FFF0F0}0.27          &\cellcolor[HTML]{FFF0F0}1.36         &\cellcolor[HTML]{FFF0F0}1.07     \\
                           & \cellcolor[HTML]{D5E3FB}Swin-VIB   &\cellcolor[HTML]{D5E3FB}\textbf{87.68}&\cellcolor[HTML]{D5E3FB}\textbf{99.48}&\cellcolor[HTML]{D5E3FB}\textbf{31.06}&\cellcolor[HTML]{F9CDCD}\textbf{0.22}&\cellcolor[HTML]{F9CDCD}\textbf{0.03}&\cellcolor[HTML]{F9CDCD}\textbf{0.54}  &\cellcolor[HTML]{D5E3FB}\textbf{62.94}&\cellcolor[HTML]{D5E3FB}96.86&\cellcolor[HTML]{D5E3FB}\textbf{27.37}&\cellcolor[HTML]{F9CDCD}\textbf{0.24}&\cellcolor[HTML]{F9CDCD}\textbf{0.16}&\cellcolor[HTML]{F9CDCD}\textbf{0.97} \\ 
                           \hline
\multirow{6}{*}{DeepSeek-7B}&\cellcolor[HTML]{F5F5F5}Closed-book&\cellcolor[HTML]{F5F5F5}16.42         &\cellcolor[HTML]{F5F5F5} -            &\cellcolor[HTML]{F5F5F5} -            &\cellcolor[HTML]{F5F5F5} 0.30           & \cellcolor[HTML]{F5F5F5}-            &\cellcolor[HTML]{F5F5F5} -            &\cellcolor[HTML]{F5F5F5}37.93         &\cellcolor[HTML]{F5F5F5}-             &\cellcolor[HTML]{F5F5F5}-             &\cellcolor[HTML]{F5F5F5} 0.33            & \cellcolor[HTML]{F5F5F5}-     &\cellcolor[HTML]{F5F5F5} -\\
                           & \cellcolor[HTML]{ECF4FF}In-context &\cellcolor[HTML]{ECF4FF}77.49         &\cellcolor[HTML]{ECF4FF}97.04         &\cellcolor[HTML]{ECF4FF}9.23          &\cellcolor[HTML]{FFF0F0}0.33          &\cellcolor[HTML]{FFF0F0}3.36          &\cellcolor[HTML]{FFF0F0}0.91          &\cellcolor[HTML]{ECF4FF}46.69         &\cellcolor[HTML]{ECF4FF}96.59         &\cellcolor[HTML]{ECF4FF}10.51       &\cellcolor[HTML]{FFF0F0}0.37          &\cellcolor[HTML]{FFF0F0}4.45          &\cellcolor[HTML]{FFF0F0}1.22   \\
                           & \cellcolor[HTML]{ECF4FF}CD$^{2}$   &\cellcolor[HTML]{ECF4FF}79.02         &\cellcolor[HTML]{ECF4FF}90.21         &\cellcolor[HTML]{ECF4FF}29.80         &\cellcolor[HTML]{FFF0F0}0.28          &\cellcolor[HTML]{FFF0F0}2.48          &\cellcolor[HTML]{FFF0F0}0.85          &\cellcolor[HTML]{ECF4FF}51.64         &\cellcolor[HTML]{ECF4FF}95.95         &\cellcolor[HTML]{ECF4FF}11.73       &\cellcolor[HTML]{FFF0F0}0.30          &\cellcolor[HTML]{FFF0F0}4.28          &\cellcolor[HTML]{FFF0F0}1.21    \\
                           & \cellcolor[HTML]{ECF4FF}Rowen-CL   &\cellcolor[HTML]{ECF4FF}81.83         &\cellcolor[HTML]{ECF4FF}96.29         &\cellcolor[HTML]{ECF4FF}50.20         &\cellcolor[HTML]{FFF0F0}0.28          &\cellcolor[HTML]{FFF0F0}2.32          &\cellcolor[HTML]{FFF0F0}0.78          &\cellcolor[HTML]{ECF4FF}43.49         &\cellcolor[HTML]{ECF4FF}94.21         &\cellcolor[HTML]{ECF4FF}9.87        &\cellcolor[HTML]{FFF0F0}0.28          &\cellcolor[HTML]{FFF0F0}4.77          &\cellcolor[HTML]{FFF0F0}1.22     \\
                           & \cellcolor[HTML]{ECF4FF}CK-PLUG    &\cellcolor[HTML]{ECF4FF}78.82         &\cellcolor[HTML]{ECF4FF}96.20         &\cellcolor[HTML]{ECF4FF}29.82         &\cellcolor[HTML]{FFF0F0}0.27          &\cellcolor[HTML]{FFF0F0}1.97          &\cellcolor[HTML]{FFF0F0}0.84         &\cellcolor[HTML]{ECF4FF}53.19         &\cellcolor[HTML]{ECF4FF}97.57         &\cellcolor[HTML]{ECF4FF}12.24       &\cellcolor[HTML]{FFF0F0}0.27          &\cellcolor[HTML]{FFF0F0}5.52         &\cellcolor[HTML]{FFF0F0}1.24     \\
                           & \cellcolor[HTML]{D5E3FB}Swin-VIB   &\cellcolor[HTML]{D5E3FB}\textbf{84.13}&\cellcolor[HTML]{D5E3FB}\textbf{98.83}&\cellcolor[HTML]{D5E3FB}\textbf{58.65}&\cellcolor[HTML]{F9CDCD}\textbf{0.24}&\cellcolor[HTML]{F9CDCD}\textbf{0.79} &\cellcolor[HTML]{F9CDCD}\textbf{0.68} &\cellcolor[HTML]{D5E3FB}\textbf{56.39}&\cellcolor[HTML]{D5E3FB}\textbf{98.75}&\cellcolor[HTML]{D5E3FB}\textbf{16.55}&\cellcolor[HTML]{F9CDCD}\textbf{0.27}&\cellcolor[HTML]{F9CDCD}\textbf{1.97} &\cellcolor[HTML]{F9CDCD}\textbf{1.10} \\ 
                           \hline
\multirow{6}{*}{Qwen3-8B}   & \cellcolor[HTML]{F5F5F5}Closed-book&\cellcolor[HTML]{F5F5F5} 23.53        &\cellcolor[HTML]{F5F5F5} -            &\cellcolor[HTML]{F5F5F5} -    &\cellcolor[HTML]{F5F5F5} 0.31              & \cellcolor[HTML]{F5F5F5}-            &\cellcolor[HTML]{F5F5F5} -            &\cellcolor[HTML]{F5F5F5}53.51         &\cellcolor[HTML]{F5F5F5}-             &\cellcolor[HTML]{F5F5F5}-             &\cellcolor[HTML]{F5F5F5} 0.34            & \cellcolor[HTML]{F5F5F5}-     &\cellcolor[HTML]{F5F5F5} -\\
                           & \cellcolor[HTML]{ECF4FF}In-context &\cellcolor[HTML]{ECF4FF}78.36         &\cellcolor[HTML]{ECF4FF}97.25         &\cellcolor[HTML]{ECF4FF}15.89         &\cellcolor[HTML]{FFF0F0}0.34          &\cellcolor[HTML]{FFF0F0}2.11          &\cellcolor[HTML]{FFF0F0}0.85          &\cellcolor[HTML]{ECF4FF}56.88         &\cellcolor[HTML]{ECF4FF}96.12         &\cellcolor[HTML]{ECF4FF}12.24       &\cellcolor[HTML]{FFF0F0}0.38          &\cellcolor[HTML]{FFF0F0}3.44         &\cellcolor[HTML]{FFF0F0}1.16   \\
                           & \cellcolor[HTML]{ECF4FF}CD$^{2}$   &\cellcolor[HTML]{ECF4FF}74.42         &\cellcolor[HTML]{ECF4FF}95.18         &\cellcolor[HTML]{ECF4FF}26.44         &\cellcolor[HTML]{FFF0F0}0.30          &\cellcolor[HTML]{FFF0F0}3.06          &\cellcolor[HTML]{FFF0F0}0.96          &\cellcolor[HTML]{ECF4FF}57.60         &\cellcolor[HTML]{ECF4FF}98.58         &\cellcolor[HTML]{ECF4FF}12.49       &\cellcolor[HTML]{FFF0F0}0.32          &\cellcolor[HTML]{FFF0F0}3.97          &\cellcolor[HTML]{FFF0F0}1.17    \\
                           & \cellcolor[HTML]{ECF4FF}Rowen-CL   &\cellcolor[HTML]{ECF4FF}80.38         &\cellcolor[HTML]{ECF4FF}97.72         &\cellcolor[HTML]{ECF4FF}21.42         &\cellcolor[HTML]{FFF0F0}0.31          &\cellcolor[HTML]{FFF0F0}1.88          &\cellcolor[HTML]{FFF0F0}0.80          &\cellcolor[HTML]{ECF4FF}54.54         &\cellcolor[HTML]{ECF4FF}97.87         &\cellcolor[HTML]{ECF4FF}14.33       &\cellcolor[HTML]{FFF0F0}0.33          &\cellcolor[HTML]{FFF0F0}4.50          &\cellcolor[HTML]{FFF0F0}1.21     \\
                           & \cellcolor[HTML]{ECF4FF}CK-PLUG    &\cellcolor[HTML]{ECF4FF}78.61         &\cellcolor[HTML]{ECF4FF}96.11         &\cellcolor[HTML]{ECF4FF}28.15         &\cellcolor[HTML]{FFF0F0}0.29          &\cellcolor[HTML]{FFF0F0}1.89          &\cellcolor[HTML]{FFF0F0}0.84          &\cellcolor[HTML]{ECF4FF}57.86         &\cellcolor[HTML]{ECF4FF}98.52         &\cellcolor[HTML]{ECF4FF}\textbf{16.78}       &\cellcolor[HTML]{FFF0F0}0.29          &\cellcolor[HTML]{FFF0F0}3.50         &\cellcolor[HTML]{FFF0F0}1.17     \\
                           & \cellcolor[HTML]{D5E3FB}Swin-VIB   &\cellcolor[HTML]{D5E3FB}\textbf{82.81}&\cellcolor[HTML]{D5E3FB}\textbf{98.68}&\cellcolor[HTML]{D5E3FB}\textbf{32.47}&\cellcolor[HTML]{F9CDCD}\textbf{0.25}&\cellcolor[HTML]{F9CDCD}\textbf{0.64} &\cellcolor[HTML]{F9CDCD}\textbf{0.82} &\cellcolor[HTML]{D5E3FB}\textbf{58.69}&\cellcolor[HTML]{D5E3FB}\textbf{99.09}&\cellcolor[HTML]{D5E3FB}15.62&\cellcolor[HTML]{F9CDCD}\textbf{0.28}&\cellcolor[HTML]{F9CDCD}\textbf{1.83}  &\cellcolor[HTML]{F9CDCD}\textbf{1.08} \\ 
                           \Xhline{1.2pt}
\end{tabular}
}
\caption{Evaluation results of multiple-choice.}
\label{table:SOTA}
\end{table*}

In this section, we verify our theoretical analysis and demonstrate that Swin-VIB can reliably guide LLMs to adapt to conflicting external context and generate the correct responses. 

\subsection{Experimental Settings}
\subsubsection{Datasets}
We use three popular datasets to evaluate the performance of Swin-VIB. Note that these datasets contain supplementary contexts and conflicting contexts, but don't have any mixed contexts.

\begin{compactitem}
  \item ConflictQA \cite{xie2023adaptive} (2,839 samples): Each $q$ contains internal knowledge "parametric memory" and conflicting external context "counter memory", mapping them to $r_s$ and $r_c$, respectively.
  \item DRUID \cite{hagstrom2024reality} (1,003 samples): Each $q$ contains a two-option item with mutually exclusive claim A/B (each paired with supporting evidence); the verified claim becomes $r_s$, the refuted one $r_c$.
  \item TruthfulQA \cite{lin2021truthfulqa} (817 samples): Each $q$ has one best-answer candidate which serves as the ground truth, multiple other correct responses, and a collection of incorrect alternatives. We use LLMs to select among the above options as $r_s$, unselected options are $r_c$. 
\end{compactitem}
Our construction only requires including queries paired with contexts from different sources. Specifically, given a query $q$, the LLM first answers, the chosen option is treated as supplementary to its internal memory, and the unchosen option as conflicting. 
For LLMs not covered by the original dataset, we follow the two-step procedure: (i) obtain a closed-book answer and rationale as Internal Memory; (ii) prompt ChatGPT to draft a coherent passage that contradicts those facts as External Conflicting Context. 
To provide the mixed contexts, a supplementary context $r_s$ and a conflicting context $r_c$ are sampled from the dataset and interleaved with each other by every four tokens, and form a mixed context $r_m$.  
\subsubsection{Models}
Our evaluations are performed on five open-source LLMs, including Llama 2-7B, Llama 2-13B, Llama 2-70B \cite{touvron2023llama}, Qwen3-8B \cite{qwen3technicalreport}, and DeepSeek-LLM-7B-base \cite{deepseek-llm}.
\begin{figure}[t]
\centering
  \includegraphics[width=0.9\columnwidth]{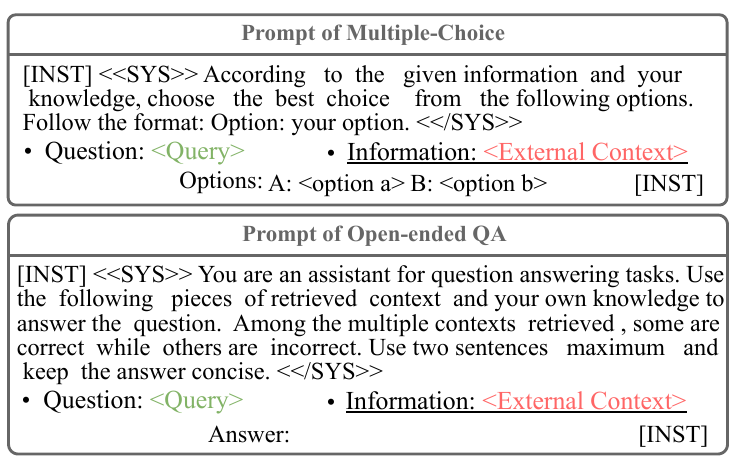}
  \caption{Prompt templates in conflicting scenarios.}
  \label{fig:experiment-prompt}
\end{figure}
\subsubsection{Tasks and Metrics}
We include two primary tasks in our experiments to evaluate the response generation of Swin-VIB when LLM retrieves conflicting contexts, and the corresponding RAG performance:\\
1) \textit{Multiple-choice}: This task can not only quantify the response generation preferences of LLMs but also can be used to evaluate the ability to exist methods to accommodate knowledge conflicts. Specifically,  we provide two options for each query, one of which is obtained from the internal memory, and the other is from the external context. Only one option is correct, which constrains the generation space and yields confirmed answers.
\begin{compactitem}
  \item Memorization Preference Rate ($\text{MPR}=\frac{f_m}{S}$);
  \item Context Preference Rate ($\text{CPR}=\frac{f_c}{S}$);
  \item Uncertain Answer Rate ($\text{UAR}=\frac{S-f_m-f_c}{S}$),
  \item Accuracy Rate($\text{ACC} = \frac{C_r}{S}$)
  \item Correction Rate ($\text{CR}=\frac{C_{crt}}{S-L}$)
  \item Resistance Rate ($\text{RR}=\frac{C_{def}}{L}$);
  \item Instance-level Uncertainty (Mean-$\psi$): Equal to the mean of the token-level negative log-likelihood values, its detailed computation procedure is provided in Appendix B.5;
  \item Total Response Entropy (TRE):\\ Equal to
        $-( ACC \log_2(ACC + \epsilon) + (1-ACC-UAR) \\
        \log_2[(1-ACC-UAR) + \epsilon] + UAR \log_2(UAR+ \epsilon))$; 
\end{compactitem}
where $S$ is the total number of dataset instances; 
$f_m$ is the number of instances for which the LLM relies on internal memory, and $f_c$ is the number of instances using external contexts. 
$L$ is the number of instances for which the LLM can generate correct responses without any context.
$C_r$ is the number of instances for which the LLM can generate correct responses with conflicting contexts.
$C_{crt}$ counts instances that are answered incorrectly without external contexts but corrected after LLM retrieved external conflicting contexts. 
$C_{def}$ counts instances that are answered correctly without external contexts and remain correctly answered after LLM retrieved conflicting contexts. 
$\epsilon$ is a small constant used to avoid numerical issues in logarithmic calculations.\\

2) \textit{Open-ended QA}: We make the LLM  generate responses, and evaluate response generation quality of RAG as well as its robustness in real-world conflicting scenarios. For each query, to evaluate an open-ended QA with retrieved context, the top five correct or incorrect answers for a query are retrieved to fill the <external context> slot in the prompt template. The template is depicted in Figure \ref{fig:experiment-prompt}. 
According to the experimental configuration of open-ended QA tasks ~\cite{yu2024truth}, we use EM and Faithfulness \cite{es2023ragas} to evaluate whether the model can provide correct answers, as well as METEOR \cite{denkowski2014meteor} to assess the quality of the response generation.

\subsubsection{Baselines and Implementation Details}
We compare Swin-VIB with five SOTA baselines (Closed-book, In-context, CD$^{2}$ \cite{jin2024tug}, CK-PLUG \cite{bi2025parameters}, Rowen-CL \cite{ding2024retrieve}) on the multiple-choice task, while integrating Swin-VIB and these baselines into three advanced RAG methods (Naive RAG, Self-RAG \cite{asai2023self}, Astute RAG \cite{wang2024astute}) for the open-ended QA task (see their implementation details in Appendix~B.4). 
We implement Swin-VIB with fixed-length sliding windows (7 tokens) and set $(\xi,\beta)$ to (0.68, $10^{-5}$) on all datasets. Data preprocessing, hardware/software environment, and model configurations are described in Appendix~B.1–B.3.
During the training stage, the bottleneck loss converges within $\approx 200$ epochs, and deeper decoder layers provide more robustness (see Appendix~C.2 for convergence analysis). 
Swin-VIB achieves advanced generation quality to $\beta$ in the range $10^{-5}\text{--}10^{-3}$, and remains stable when the threshold $\xi$ lies between 0.60 and 0.8. A window length of 7 tokens can trade off accuracy and inference latency. The windows with fewer tokens degrade the precision to accept these windows, whereas the windows with more than 7 tokens also slightly degrade acceptance precision. The parameters $\beta$ and $\xi$, and the window length are analyzed in Appendix C.3.

\subsection{Verification of Our Theoretical Analysis}
\label{sec-II}
\begin{figure}[t]
\centering
  \includegraphics[width=0.85\columnwidth]{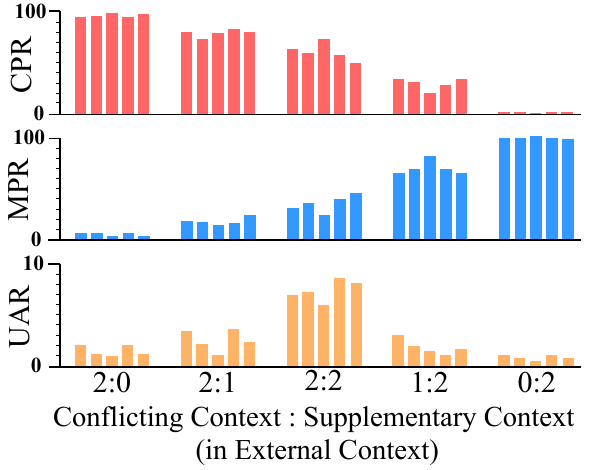}
  \caption{The uncertainty answer ratio of LLMs under varying proportions of external conflicting information from ConflictQA, e.g., 1:2, means that the external context includes one conflicting context and two supplementary contexts. For each bar group, the results for different LLMs are listed according to the order of Llama2-7B, Llama2-13B, Llama2-70B, Deepseek-7B and Qwen3-8B.}
  \label{fig:proportion}
\end{figure}

To verify our theoretical analysis~\ref{sec:Theoretical_Analysis}, we explore the impact of information difference on LLMs' preference and generation uncertainty on the multiple-choice task. To achieve that, the internal-memory facts (supplementary information) are collected and combined into the external context with controlled proportions. As depicted in Figure \ref{fig:detecting: proportion}, we get two observations:
\begin{compactitem}
  \item The preferences of the LLM are influenced by the proportion of external conflicting and supplementary information.  As the information difference grows, LLM becomes more focused on a certain preference, and the curve in  Figure~\ref{fig:detecting: proportion} mirrors this trajectory. (see Appendix C.1 for analysis of answer preference distributions)
  \item When the proportions are set at 2:2, the UAR values reach their maximum. This indicates that when the information difference is slight, the uncertainty of response generation is the greatest. The empirical “rise–then–fall” trend of the UAR in Figure~\ref{fig:proportion} is consistent with the curve in Figure~\ref{fig:detecting: proportion}. 
\end{compactitem}
\subsection{Comparisons with State of the Arts}
Through the comparison of Swin-VIB and baselines, we can observe the following in Table \ref{table:SOTA}:
Swin-VIB significantly outperforms the state-of-the-art baselines on all settings and outperforms the strongest baseline by up to 6.24\%. This demonstrates its robustness of response generation in the conflicting scenario. Swin-VIB attains the best trade-off between CR and RR, reflecting both strong capability for error recovery and resilience to misleading context. This enables Swin-VIB to generate more reliable responses.
On the other hand, it considerably reduces UAR, so the LLMs have less probability of skipping answering the queries. To specifically measure the uncertainty of response generation, following the literature \cite{zhu2024unraveling}, we take the output distribution of the LLM to measure the macro-level uncertainty of LLMs with the metric, total response entropy (TRE). The corresponding results prove that the adaptation principle of Swin-VIB is quite effective in accommodating knowledge conflicts and improving response reliability. 
Similarly, the decrease in Mean-$\psi$ confirms that Swin-VIB maintains smaller micro-level uncertainty. Actually, the Pearson correlation between $Mean-\psi$ and TRE is 0.81 (see Appendix C.5), validating the high robustness of Swin-VIB. Finally, to verify whether Swin-VIB has selected the windows with larger information difference, we conducted a small-scale human annotation study. We find that 76 \% of the windows discarded by Swin-VIB were judged “undecidable” (no clear conflicting or supplementary information) and the result confirms that our method indeed amplifies the information difference in a rational way (see Appendix C.6).

\subsection{Evaluation of RAG with Swin-VIB}

This section examines how integrating Swin-VIB into advanced RAG frameworks, enhances practical applicability under real-world settings, including evaluating generation quality and inference latency.

\begin{table}[ht]
\renewcommand\arraystretch{1}
\footnotesize
\setlength{\tabcolsep}{1.2mm}{
\begin{tabular}{lccc}
\Xhline{1.2pt}
Methods               & EM$\uparrow$ &Meteor$\uparrow$&Faithfulness $\downarrow$ \\ \hline
Naive RAG             &\cellcolor[HTML]{ECF4FF}46.50     &\cellcolor[HTML]{ECF4FF}42.85        &\cellcolor[HTML]{FFF0F0}73.03             \\
Naive RAG + Swin-VIB  &\cellcolor[HTML]{D5E3FB}60.14     &\cellcolor[HTML]{D5E3FB}50.95        &\cellcolor[HTML]{F9CDCD}66.58                                     \\ \hline
Self-RAG              &\cellcolor[HTML]{ECF4FF}39.39     &\cellcolor[HTML]{ECF4FF}43.38            &\cellcolor[HTML]{FFF0F0}70.10              \\
Self-RAG + Swin-VIB   &\cellcolor[HTML]{D5E3FB}58.32     &\cellcolor[HTML]{D5E3FB}55.29       &\cellcolor[HTML]{F9CDCD}64.32              \\ \hline
Astute RAG            &\cellcolor[HTML]{ECF4FF}53.02     &\cellcolor[HTML]{ECF4FF}44.69        &\cellcolor[HTML]{FFF0F0}65.23              \\
Astute RAG + Swin-VIB &\cellcolor[HTML]{D5E3FB}64.16     &\cellcolor[HTML]{D5E3FB}53.70        &\cellcolor[HTML]{F9CDCD}64.09             \\ \Xhline{1.2pt}
\end{tabular}}
\caption{Evaluation of Open-ended QA tasks on TruthfulQA}
\label{table:rag}
\end{table}

As shown in Table \ref{table:rag}, Astute-RAG employs an iterative fusion step in a data augmented stage, and thus raises decision variance and generation uncertainty. From the beginning, Swin-VIB drops the windows with high uncertainty context, alleviates Astute-RAG’s uncertainty, and achieves an improvement of EM by 11.14\%. Additionally, we observe a significant reduction in faithfulness by 6.45\%, indicating that the LLM no longer copies the retrieved context totally. This is because the sliding-window strategy of Swin-VIB trades some cross-segment coherence for higher intra-segment consistency. 
On the other hand, the naive RAG and Self-RAG lack conflicting accommodation resolution, often copy the retrieved context, and fall behind on both EM and METEOR. The high EM scores of Swin-VIB show that it can adapt the LLM’s preference on external contexts and select the most relevant external information, even when the top-5 retrieved contexts contain conflicting information. 
These results demonstrate that Swin-VIB can improve response generation quality and reliability for mitigating knowledge conflicts in RAG.

\begin{table}[ht]
 \renewcommand\arraystretch{1}
 \footnotesize
 \setlength{\tabcolsep}{1mm}{
\begin{tabular}{lccc}
\Xhline{1.2pt}
LLM                  & Naive RAG            & Our cost per window$\downarrow$            & Our cost$\downarrow$            \\ \hline
Llama2-7B            &\cellcolor[HTML]{FFF0F0}0.4912s     &\cellcolor[HTML]{F9CDCD}0.08ms            &\cellcolor[HTML]{F9CDCD}0.3913s                \\ 
Llama2-13B           &\cellcolor[HTML]{FFF0F0}0.8791s     &\cellcolor[HTML]{F9CDCD}0.14ms            &\cellcolor[HTML]{F9CDCD}0.5221s            \\ 
Llama2-70B           &\cellcolor[HTML]{FFF0F0}4.6013s     &\cellcolor[HTML]{F9CDCD}0.55ms            &\cellcolor[HTML]{F9CDCD}2.7830s                \\ 
Qwen3-8B              &\cellcolor[HTML]{FFF0F0}0.3862s     &\cellcolor[HTML]{F9CDCD}0.12ms            &\cellcolor[HTML]{F9CDCD}0.2740s                 \\ 
DeepSeek-7B          &\cellcolor[HTML]{FFF0F0}0.3770s     &\cellcolor[HTML]{F9CDCD}0.11ms            &\cellcolor[HTML]{F9CDCD}0.4293s          \\ \Xhline{1.2pt}
\end{tabular}}
 \caption{Efficiency Evaluation of Open-ended QA tasks on TruthfulQA}
 \label{table:time}
\end{table}
Table \ref{table:time} evaluates the overhead of Swin-VIB in response generation delay. For Llama2-7B, Swin-VIB involves an additional delay of 0.3913 seconds. It can also be found that after the Swin-VIB has been integrated into the RAG systems, the increased number of model parameters does not significantly increase the latency. The Rag systems remain sufficiently lightweight, since Swin-VIB trades off efficiency and accuracy in response generation. (see Appendix C.4 for computational complexity analysis).

\section{Related Work}
Efforts to mitigate risks associated with knowledge conflicts in response generation can be categorized into three groups: a) Internal knowledge-driven methods \cite{shi2023context, jin2024cutting, zhang2024truthx}: Fine-tuning or editing the model so that the retrieved evidences can override conflicts; b) External-validation methods incorporate verification module to ensure the reliability of retrieved contexts \cite{yu2024truth, kortukov2024studying, bi2025parameters}; c) Adaptive methods at the decoding stage, take LLM parameters frozen and detect token-level conflicts through adjusting parametric and context logits \cite{shi2023trusting, yuan2024discerning, jin2024tug, bi2025parameters, huang2025pip}. However, these existing methods have limitations in elucidating LLMs' knowledge preference due to: a) Internal catastrophic forgetting of LLMs caused by fine-tuning~\cite{xu2024knowledge}, b) Overconfidence in external sources caused by external validation~\cite {xie2023adaptive}; c) Adaptive single-token adjustments struggling with long text analysis, thus limiting their practical applicability in real-world RAG deployments. The recent empirical works \cite{jin2024cutting, zhao2024steering, yuan2025exploiting} focus on heuristic interventions on response generation of LLMs. 

In contrast, our theoretical study demonstrates that the information difference between external context and internal representations is the true force to drive preference shifts and information selection. According to this insight, we propose a novel response generation method \cite{li2024matching}, Swin-VIB. Its bottleneck-based adapter is involved to regulate the compression ratio of the retrieved information, without retraining the pre-trained LLMs. 
In this way, Swin-VIB accommodates conflicting knowledge, avoiding merely relying on internal or external information. 
This information-theoretic insight furnishes a unified, \red{model-agnostic yardstick that combines theoretical breadth with practical simplicity.}

\section{Conclusion}
This work proposes a novel theoretical framework to analyze and address the issue of information conflicts encountered by LLMs in RAG systems. Leveraging analysis on the knowledge conflicts and preferences of LLMs from an information theory perspective, we find that the uncertainty of LLMs can be mitigated by adapting the external information difference of LLMs. Insight from this, we propose Swin-VIB to optimize how RAG handles external contexts to achieve reliable response generation. Experimental results demonstrate that Swin-VIB significantly accommodates conflicts, reduces uncertainty in LLM outputs, and generates more accurate, consistent, and context-aware responses. Moreover, Swin-VIB enhances retrieval system performance, facilitating its real-world applications. Future work will explore extending this approach to more types of response generation tasks to validate further and to improve its effectiveness.

\section*{Acknowledgments}
This work is partially supported by the National Natural Science Foundation of China (62272248 and 61962045), the Natural Science Foundation of Tianjin (25JJJJC0016, 23JCZDJC01010 and 25JCZDSN00040), in part by the Program for Young Talents of Science and Technology in Universities of Inner Mongolia Autonomous Region under Grant NJYT23104, and the Basic Scientific Research Expenses Program of Universities directly under Inner Mongolia Autonomous Region under Grant JY20220273 and Grant JY20240002.

\begin{thebibliography}{40}
\providecommand{\natexlab}[1]{#1}

\bibitem[{Alemi et~al.(2016)Alemi, Fischer, Dillon, and Murphy}]{alemi2016deep}
Alemi, A.~A.; Fischer, I.; Dillon, J.~V.; and Murphy, K. 2016.
\newblock Deep variational information bottleneck.
\newblock \emph{arXiv preprint arXiv:1612.00410}.

\bibitem[{Asai et~al.(2023)Asai, Wu, Wang, Sil, and Hajishirzi}]{asai2023self}
Asai, A.; Wu, Z.; Wang, Y.; Sil, A.; and Hajishirzi, H. 2023.
\newblock Self-rag: Learning to retrieve, generate, and critique through self-reflection.
\newblock \emph{arXiv preprint arXiv:2310.11511}.

\bibitem[{Bi et~al.(2025)Bi, Liu, Wang, Xu, Fang, Mei, and Cheng}]{bi2025parameters}
Bi, B.; Liu, S.; Wang, Y.; Xu, Y.; Fang, J.; Mei, L.; and Cheng, X. 2025.
\newblock Parameters vs. Context: Fine-Grained Control of Knowledge Reliance in Language Models.
\newblock \emph{arXiv preprint arXiv:2503.15888}.

\bibitem[{Brown et~al.(2020)Brown, Mann, Ryder, Subbiah, Kaplan, Dhariwal, Neelakantan, Shyam, Sastry, Askell et~al.}]{brown2020language}
Brown, T.; Mann, B.; Ryder, N.; Subbiah, M.; Kaplan, J.~D.; Dhariwal, P.; Neelakantan, A.; Shyam, P.; Sastry, G.; Askell, A.; et~al. 2020.
\newblock Language models are few-shot learners.
\newblock \emph{Advances in neural information processing systems}, 33: 1877--1901.

\bibitem[{DeepSeek-AI(2024)}]{deepseek-llm}
DeepSeek-AI. 2024.
\newblock DeepSeek LLM: Scaling Open-Source Language Models with Longtermism.
\newblock \emph{arXiv preprint arXiv:2401.02954}.

\bibitem[{Denkowski and Lavie(2014)}]{denkowski2014meteor}
Denkowski, M.; and Lavie, A. 2014.
\newblock Meteor universal: Language specific translation evaluation for any target language.
\newblock In \emph{Proceedings of the ninth workshop on statistical machine translation}, 376--380.

\bibitem[{Ding et~al.(2024)Ding, Pang, Wei, Shen, and Cheng}]{ding2024retrieve}
Ding, H.; Pang, L.; Wei, Z.; Shen, H.; and Cheng, X. 2024.
\newblock Retrieve only when it needs: Adaptive retrieval augmentation for hallucination mitigation in large language models.
\newblock \emph{arXiv preprint arXiv:2402.10612}.

\bibitem[{Elastic(2023)}]{elasticsearch}
Elastic. 2023.
\newblock Elasticsearch.
\newblock \url{https://www.elastic.co/guide/en/elasticsearch/reference/current/elasticsearch-intro-what-is-es.html}.
\newblock Accessed: 2023-10-10.

\bibitem[{Es et~al.(2023)Es, James, Espinosa-Anke, and Schockaert}]{es2023ragas}
Es, S.; James, J.; Espinosa-Anke, L.; and Schockaert, S. 2023.
\newblock Ragas: Automated evaluation of retrieval augmented generation.
\newblock \emph{arXiv preprint arXiv:2309.15217}.

\bibitem[{Gao et~al.(2023)Gao, Xiong, Gao, Jia, Pan, Bi, Dai, Sun, Wang, and Wang}]{gao2023retrieval}
Gao, Y.; Xiong, Y.; Gao, X.; Jia, K.; Pan, J.; Bi, Y.; Dai, Y.; Sun, J.; Wang, H.; and Wang, H. 2023.
\newblock Retrieval-augmented generation for large language models: A survey.
\newblock \emph{arXiv preprint arXiv:2312.10997}, 2: 1.

\bibitem[{Hagstr{\"o}m et~al.(2024)Hagstr{\"o}m, Marjanovi{\'c}, Yu, Arora, Lioma, Maistro, Atanasova, and Augenstein}]{hagstrom2024reality}
Hagstr{\"o}m, L.; Marjanovi{\'c}, S.~V.; Yu, H.; Arora, A.; Lioma, C.; Maistro, M.; Atanasova, P.; and Augenstein, I. 2024.
\newblock A Reality Check on Context Utilisation for Retrieval-Augmented Generation.
\newblock \emph{arXiv preprint arXiv:2412.17031}.

\bibitem[{Huang et~al.(2025)Huang, Liu, Yan, Yi, Chen, Liu, Sun, Xiao, Yu, and Xiong}]{huang2025pip}
Huang, P.; Liu, Z.; Yan, Y.; Yi, X.; Chen, H.; Liu, Z.; Sun, M.; Xiao, T.; Yu, G.; and Xiong, C. 2025.
\newblock Pip-kag: Mitigating knowledge conflicts in knowledge-augmented generation via parametric pruning.
\newblock \emph{arXiv preprint arXiv:2502.15543}.

\bibitem[{Jin et~al.(2024{\natexlab{a}})Jin, Cao, Chen, Liu, Jiang, Xu, Li, and Zhao}]{jin2024tug}
Jin, Z.; Cao, P.; Chen, Y.; Liu, K.; Jiang, X.; Xu, J.; Li, Q.; and Zhao, J. 2024{\natexlab{a}}.
\newblock Tug-of-war between knowledge: Exploring and resolving knowledge conflicts in retrieval-augmented language models.
\newblock \emph{arXiv preprint arXiv:2402.14409}.

\bibitem[{Jin et~al.(2024{\natexlab{b}})Jin, Cao, Yuan, Chen, Xu, Li, Jiang, Liu, and Zhao}]{jin2024cutting}
Jin, Z.; Cao, P.; Yuan, H.; Chen, Y.; Xu, J.; Li, H.; Jiang, X.; Liu, K.; and Zhao, J. 2024{\natexlab{b}}.
\newblock Cutting Off the Head Ends the Conflict: A Mechanism for Interpreting and Mitigating Knowledge Conflicts in Language Models.
\newblock \emph{arXiv preprint arXiv:2402.18154}.

\bibitem[{Kingma and Welling(2013)}]{kingma2013auto}
Kingma, D.~P.; and Welling, M. 2013.
\newblock Auto-encoding variational bayes.
\newblock \emph{arXiv preprint arXiv:1312.6114}.

\bibitem[{Kong and Kim(2024)}]{kong2024posterior}
Kong, I.; and Kim, Y. 2024.
\newblock Posterior concentrations of fully-connected Bayesian neural networks with general priors on the weights.
\newblock \emph{arXiv preprint arXiv:2403.14225}.

\bibitem[{Kortukov et~al.(2024)Kortukov, Rubinstein, Nguyen, and Oh}]{kortukov2024studying}
Kortukov, E.; Rubinstein, A.; Nguyen, E.; and Oh, S.~J. 2024.
\newblock Studying large language model behaviors under realistic knowledge conflicts.
\newblock \emph{arXiv e-prints}, arXiv--2404.

\bibitem[{Lewis et~al.(2020)Lewis, Perez, Piktus, Petroni, Karpukhin, Goyal, K{\"u}ttler, Lewis, Yih, Rockt{\"a}schel et~al.}]{lewis2020retrieval}
Lewis, P.; Perez, E.; Piktus, A.; Petroni, F.; Karpukhin, V.; Goyal, N.; K{\"u}ttler, H.; Lewis, M.; Yih, W.-t.; Rockt{\"a}schel, T.; et~al. 2020.
\newblock Retrieval-augmented generation for knowledge-intensive nlp tasks.
\newblock \emph{Advances in Neural Information Processing Systems}, 33: 9459--9474.

\bibitem[{Li et~al.(2024)Li, Jin, Zhou, Zhang, Zhang, Zhu, and Dou}]{li2024matching}
Li, X.; Jin, J.; Zhou, Y.; Zhang, Y.; Zhang, P.; Zhu, Y.; and Dou, Z. 2024.
\newblock From matching to generation: A survey on generative information retrieval.
\newblock \emph{arXiv preprint arXiv:2404.14851}.

\bibitem[{Lin, Hilton, and Evans(2021)}]{lin2021truthfulqa}
Lin, S.; Hilton, J.; and Evans, O. 2021.
\newblock Truthfulqa: Measuring how models mimic human falsehoods.
\newblock \emph{arXiv preprint arXiv:2109.07958}.

\bibitem[{{OpenAI}(2023)}]{openai2023gpt}
{OpenAI}. 2023.
\newblock GPT-4 Technical Report.
\newblock \emph{arXiv preprint arXiv:2303.08774}.

\bibitem[{Polson and Ro{\v{c}}kov{\'a}(2018)}]{polson2018posterior}
Polson, N.~G.; and Ro{\v{c}}kov{\'a}, V. 2018.
\newblock Posterior concentration for sparse deep learning.
\newblock \emph{Advances in neural information processing systems}, 31.

\bibitem[{Robertson and Jones(1976)}]{robertson1976relevance}
Robertson, S.~E.; and Jones, K.~S. 1976.
\newblock Relevance weighting of search terms.
\newblock \emph{Journal of the American Society for Information science}, 27(3): 129--146.

\bibitem[{Shannon(1948)}]{shannon1948mathematical}
Shannon, C.~E. 1948.
\newblock A mathematical theory of communication.
\newblock \emph{The Bell system technical journal}, 27(3): 379--423.

\bibitem[{Shi et~al.(2023{\natexlab{a}})Shi, Han, Lewis, Tsvetkov, Zettlemoyer, and Yih}]{shi2023trusting}
Shi, W.; Han, X.; Lewis, M.; Tsvetkov, Y.; Zettlemoyer, L.; and Yih, S. W.-t. 2023{\natexlab{a}}.
\newblock Trusting your evidence: Hallucinate less with context-aware decoding.
\newblock \emph{arXiv preprint arXiv:2305.14739}.

\bibitem[{Shi et~al.(2023{\natexlab{b}})Shi, Min, Lomeli, Zhou, Li, Lin, Smith, Zettlemoyer, Yih, and Lewis}]{shi2023context}
Shi, W.; Min, S.; Lomeli, M.; Zhou, C.; Li, M.; Lin, V.; Smith, N.~A.; Zettlemoyer, L.; Yih, S.; and Lewis, M. 2023{\natexlab{b}}.
\newblock In-context pretraining: Language modeling beyond document boundaries.
\newblock \emph{arXiv preprint arXiv:2310.10638}.

\bibitem[{Team(2025)}]{qwen3technicalreport}
Team, Q. 2025.
\newblock Qwen3 Technical Report.
\newblock arXiv:2505.09388.

\bibitem[{Tishby, Pereira, and Bialek(2000)}]{tishby2000information}
Tishby, N.; Pereira, F.~C.; and Bialek, W. 2000.
\newblock The information bottleneck method.
\newblock \emph{arXiv preprint physics/0004057}.

\bibitem[{Touvron et~al.(2023)Touvron, Martin, Stone, Albert, Almahairi, Babaei, Bashlykov, Batra, Bhargava, Bhosale et~al.}]{touvron2023llama}
Touvron, H.; Martin, L.; Stone, K.; Albert, P.; Almahairi, A.; Babaei, Y.; Bashlykov, N.; Batra, S.; Bhargava, P.; Bhosale, S.; et~al. 2023.
\newblock Llama 2: Open foundation and fine-tuned chat models.
\newblock \emph{arXiv preprint arXiv:2307.09288}.

\bibitem[{Wang et~al.(2024)Wang, Wan, Sun, Chen, and Ar{\i}k}]{wang2024astute}
Wang, F.; Wan, X.; Sun, R.; Chen, J.; and Ar{\i}k, S.~{\"O}. 2024.
\newblock Astute rag: Overcoming imperfect retrieval augmentation and knowledge conflicts for large language models.
\newblock \emph{arXiv preprint arXiv:2410.07176}.

\bibitem[{Wang, Sun, and He(2023)}]{M3E}
Wang, Y.; Sun, Q.; and He, S. 2023.
\newblock M3E: Moka Massive Mixed Embedding Model.
\newblock Online.
\newblock Accessed: 2025-11-15.

\bibitem[{Wu, Wu, and Zou(2024)}]{wu2024faithful}
Wu, K.; Wu, E.; and Zou, J. 2024.
\newblock How faithful are RAG models? Quantifying the tug-of-war between RAG and LLMs' internal prior.
\newblock \emph{arXiv preprint arXiv:2404.10198}.

\bibitem[{Xie et~al.(2023)Xie, Zhang, Chen, Lou, and Su}]{xie2023adaptive}
Xie, J.; Zhang, K.; Chen, J.; Lou, R.; and Su, Y. 2023.
\newblock Adaptive chameleon or stubborn sloth: Revealing the behavior of large language models in knowledge conflicts.
\newblock In \emph{The Twelfth International Conference on Learning Representations}.

\bibitem[{Xu et~al.(2024)Xu, Qi, Wang, Wang, Zhang, and Xu}]{xu2024knowledge}
Xu, R.; Qi, Z.; Wang, C.; Wang, H.; Zhang, Y.; and Xu, W. 2024.
\newblock Knowledge Conflicts for LLMs: A Survey.
\newblock \emph{arXiv preprint arXiv:2403.08319}.

\bibitem[{Yu, Zhang, and Feng(2024)}]{yu2024truth}
Yu, T.; Zhang, S.; and Feng, Y. 2024.
\newblock Truth-aware context selection: Mitigating hallucinations of large language models being misled by untruthful contexts.
\newblock \emph{arXiv preprint arXiv:2403.07556}.

\bibitem[{Yuan et~al.(2025)Yuan, Yang, Huang, Wang, Fan, Ju, Zhao, and Liu}]{yuan2025exploiting}
Yuan, X.; Yang, Z.; Huang, Z.; Wang, Y.; Fan, S.; Ju, Y.; Zhao, J.; and Liu, K. 2025.
\newblock Exploiting Contextual Knowledge in LLMs through V-usable Information based Layer Enhancement.
\newblock \emph{arXiv preprint arXiv:2504.15630}.

\bibitem[{Yuan et~al.(2024)Yuan, Yang, Wang, Liu, Zhao, and Liu}]{yuan2024discerning}
Yuan, X.; Yang, Z.; Wang, Y.; Liu, S.; Zhao, J.; and Liu, K. 2024.
\newblock Discerning and resolving knowledge conflicts through adaptive decoding with contextual information-entropy constraint.
\newblock \emph{arXiv preprint arXiv:2402.11893}.

\bibitem[{Zhang, Yu, and Feng(2024)}]{zhang2024truthx}
Zhang, S.; Yu, T.; and Feng, Y. 2024.
\newblock Truthx: Alleviating hallucinations by editing large language models in truthful space.
\newblock \emph{arXiv preprint arXiv:2402.17811}.

\bibitem[{Zhao et~al.(2024)Zhao, Devoto, Hong, Du, Gema, Wang, He, Wong, and Minervini}]{zhao2024steering}
Zhao, Y.; Devoto, A.; Hong, G.; Du, X.; Gema, A.~P.; Wang, H.; He, X.; Wong, K.-F.; and Minervini, P. 2024.
\newblock Steering knowledge selection behaviours in LLMs via sae-based representation engineering.
\newblock \emph{arXiv preprint arXiv:2410.15999}.

\bibitem[{Zhu et~al.(2024)Zhu, Liu, Wang, Tu, and Chen}]{zhu2024unraveling}
Zhu, T.; Liu, Q.; Wang, F.; Tu, Z.; and Chen, M. 2024.
\newblock Unraveling cross-modality knowledge conflicts in large vision-language models.
\newblock \emph{arXiv preprint arXiv:2410.03659}.

\end{thebibliography}

\clearpage
\appendix
\setcounter{secnumdepth}{2}
\setcounter{section}{0}
\twocolumn[%
\begin{center}
\LARGE\bfseries
Technical Appendix for Accommodate Knowledge Conflicts in Retrieval-augmented LLMs: Towards Robust Response Generation in the Wild
\par
\vspace{5em}
\normalsize
\end{center}
\vspace{1em}
]

\section{Theoretical Analysis}
\subsection{Factorising Conditional Entropy}
\label{sec:appendix:symbol_definition}
Without a theoretical analysis of knowledge conflicts, the limitations of empirical rules and experimental settings in response generation can hardly be alleviated \cite {xu2024knowledge, wu2024faithful, xie2023adaptive, jin2024tug}. 
To tackle this challenge, we propose a theoretical framework that shows knowledge conflicts can be defined by conditional entropy, the corresponding symbol system is defined in Table \ref{tab:notion1}.
\begin{table}[h]
\renewcommand\arraystretch{1}
\centering
\caption{Symbols and their meanings}
\label{tab:notion1}
\setlength{\tabcolsep}{1mm}{
\begin{tabular}{ll}
\hline
Symbol        & Meaning                                                                                                                                                            \\ \hline
$Q$           & \begin{tabular}[c]{@{}l@{}}Queries\end{tabular}                                                  \\ \hline
$R=B(Q)$      & \begin{tabular}[c]{@{}l@{}}External contexts from the \\ knowledge base $B$\end{tabular} \\ \hline
$O=LLM(R, Q)$ & \begin{tabular}[c]{@{}l@{}}Generated Responses by LLM\end{tabular}                  \\ \hline
\end{tabular}}
\end{table}
Based on Definition 1 (Uncertainty, main paper), we derive from the law of total probability and the chain rule: 
\begin{equation}
\begin{aligned}
\mathbb{H}(O\!\mid\!R,Q)
&= -\sum_{q,r,o} p(q)\,p(r\mid q)\,p(o\mid r,q)\log p(o\mid r,q) \\
&= \sum_{q} p(q)\sum_{r} p(r\mid q)\sum_{o}\psi\!\bigl(p(o\mid r,q)\bigr).
\end{aligned}
\end{equation}
Let $\psi(\cdot)$ denote instance-level uncertainty, and be calculated by
\begin{equation}
\psi(p(o\mid r, q))=-p(o\mid r, q)\log p(o\mid r, q)
\end{equation}

\subsection{Decomposing Conflicting and Supplementary Information}
\label{sec:appendix:derivation}
$p(o \mid r, q)$ can be derived as the follows:
\begin{equation}
\label{fm:p(}
\begin{aligned}
&p(o \mid r, q)\\
& = \int_{X} p\left(o \mid r, q, x\right) p\left(x\mid r,q\right) dx, \\
& \quad \text{\qquad//marginalization} \\
& = \int_{X} p\left(o \mid r, q, x\right) \frac{p\left(r,q\mid x\right)p(x)}{p\left(r, q\right)} dx ,\\
& \quad \text{\qquad//Bayes' Theorem} \\
& \propto \int_{X} p\left(o \mid r, q, x\right) p\left(r,q\mid x\right)p(x) dx, \\
& \quad \text{\qquad//$p\left(r, q\right)$ is a constant}\\
& \propto \int_{X} p\left(o \mid r, q, x\right) \frac{p\left(r,q\mid x\right)}{p\left(r,q\mid x_{\gamma}\right)} p(x)dx,\\ 
& \quad  \text{\qquad//$p\left(r,q\mid x_{\gamma}\right)$ is a constant}\\
& = \int_{X} p\left(o \mid r, q, x\right) exp\Biggl[\log \frac{p\left(r,q\mid x\right)}{p\left(r,q\mid x_{\gamma}\right)} \Biggl]p(x)dx\\
&= \int_X p(o\mid r,q,x)\,
\exp\!\Biggl[
\underbrace{\log\frac{p(q\mid r,x)}{p(q\mid r,x_\gamma)}}_{\delta_q(r,x)}+\\
&\qquad\qquad\qquad\qquad\qquad\log\frac{p(r\mid x)}{p(r\mid x_\gamma)}\Biggr] p(x)\,dx.\\
& \approx \int_{X} p\left(o \mid r, q, x\right) exp\Biggl[\log \mathcal{B} +\\
& \qquad\qquad\qquad\qquad\qquad  \log \frac{p(r \mid x)}{p\left(r \mid x_{\gamma}\right)} \Biggl]p(x)dx\\
& \approx \int_{X} p\left(o \mid r, q, x\right) exp\Biggl[\log{p\left(r \mid x\right)}-\\
& \qquad\qquad\qquad\qquad\qquad \log{p\left(r \mid x_{\gamma}\right)} \Biggl]p(x)dx\\
\end{aligned}
\end{equation}
where $X$ denotes the high-dimensional latent space of the LLM containing all embedded information that may affect response generation; $x$ is a recalled instance sampled from $X$. Let $\delta_q(r,x)=\log\frac{p(q\mid r,x)}{p(q\mid r,x_\gamma)}$, 
and $s(x)=\log p(r,q\mid x)$ be the retriever score and $x_{\gamma}$ is a top-ranked latent state that coincides with information retrieved from an external text or database, while considering query $q$. So $s(x_\gamma)\ge s(x)$ and
\begin{equation}
\log\frac{p(r,q\mid x_\gamma)}{p(r,q\mid x)}\ge 0.
\end{equation}
Using Bayes’ rule, this fraction can be divded:
\begin{equation}
\begin{aligned}
\delta_q(r,x)
&=\log\frac{p(r,q\mid x)}{p(r,q\mid x_\gamma)}+\log\frac{p(r\mid x_\gamma)}{p(r\mid x)}\\
&\;\le\;
\log\frac{p(r\mid x_\gamma)}{p(r\mid x)}
=\log\mathcal B(r,x),
\end{aligned}
\end{equation}
which keeps $e^{\delta_q(r,x)}$ uniformly bounded on the candidate set. Around $x_\gamma$, a first-order expansion gives
\begin{equation}
e^{\delta_q(r,x)} = e^{\delta_q(r,x_\gamma)}\,[\,1+\mathcal O(\|x-x_\gamma\|)\,].
\end{equation}
The factor $e^{\delta_q(r,x_\gamma)}$ is independent of $o$, and the residual $\mathcal O(\|x-x_\gamma\|)$ is small to the integral. Therefore, in
\begin{equation}
\begin{aligned}
p(o\mid r,q)\;=\;\int_X p(o\mid r,q,x)\,&e^{\delta_q(r,x_\gamma)}\,[\,1+\mathcal O(\|x-x_\gamma\|)\,]\,\\
&\exp\!\Bigl[\log\tfrac{p(r\mid x)}{p(r\mid x_\gamma)}\Bigr]\,dx,
\end{aligned}
\end{equation}
Any factor that does not depend on $o$ can be absorbed by normalization (equivalently, by normalizing the weights over $x$). Therefore we use
\begin{equation}
p(o\mid r,q)
\propto \int_X p(o\mid r,q,x)\,
\exp\!\Bigl[\log\tfrac{p(r\mid x)}{p(r\mid x_\gamma)}\Bigr] p(x)\,dx .
\label{eq:a2-main}
\end{equation}
and drop $\delta_q$ thereafter since it only induces a first-order (small) perturbation of the weights.

\subsection{First-Order Approximation}
The self-information of $-\log p(\cdot)$ quantifies how much information is revealed~\cite{shannon1948mathematical}. We categorize the external information retrieved from an external text or database into two types, whereas the LLMs' pre-trained knowledge is either contradicted or insufficient. 
\begin{compactitem}
    \item \textit{Conflicting Information}: If the retrieved information instance contradicts $x_{\gamma}$ in the LLMs' pre-trained knowledge, quantified by \\$I_c=-\log{p\left(r \mid x_{\gamma}\right)}$.
    \item \textit{Supplementary Information}: If the retrieved information is complementary to the external context and quantified $I_s=-\log{p\left(r \mid x \right)}$.
\end{compactitem}
According to Formula~\ref{eq:a2-main}, we write
\begin{equation}
\label{eq:A1}
\begin{aligned}
&p(o\mid r,q) \;\propto\
\int_{X} p(o\mid r,q,x)\,
       \exp\!\bigl[I_c - I_s\bigr]\,
       p(x)\,dx .
\end{aligned}
\end{equation}
From Eq.~\eqref{eq:A1} we have
\begin{equation}
p(o\mid r,q)\propto e^{I_c}\!\int_X p(o\mid r,q,x)\,e^{-I_s(x)}\,p(x)\,dx,
\tag{A.8}
\end{equation}
Let $w(x)=p(o\mid r,q,x)\,e^{-I_s(x)}\,p(x)$, $\phi(x)=-\log w(x)$.
Because $p(x\mid r,q)$ for LLMs is sharply peaked \cite{polson2018posterior,kong2024posterior}, $w(x)$ is concentrated in a narrow region around a local minimum $x^\star=\arg\min_x \phi(x)$ (equivalently, a maximum of $w$). 
A second-order Taylor expansion of $\phi$ around $x^\star$ yields
\begin{equation}
\phi(x)\approx \phi(x^\star)+\tfrac12 (x-x^\star)^\top H(x^\star)(x-x^\star),
\end{equation}
where $H$ is the Hessian. Evaluating the Gaussian integral gives
\begin{equation}
p(o\mid r,q)\approx \tilde C(o)\,\exp\!\big[I_c - I_s(x^\star)\big] + \mathcal O(\varepsilon^2),
\tag{A.9}
\end{equation}
where $\tilde C(o)$ collects terms independent of $x$ (and $\varepsilon$ characterises the local width). After normalising over $o$, constants such as $\tilde C(o)$ cancel, so the relative probability is proportional to $\exp\!\big[I_c - I_s(x^\star)\big]$. Since $\exp$ is strictly increasing,
\begin{equation}
\Delta I = I_c - I_s(x^\star).
\end{equation}
We use the first-order linear proxy
\begin{equation}
\exp(\Delta I)\approx 1+\Delta I,
\end{equation}
hence $p(o\mid r,q)\propto \Delta I$. Larger $|\Delta I|$ pushes the LLM towards the source (conflicting information vs.\ supplementary information) that better explains the retrieved context, thereby increasing the relative weight of answers consistent with that source.

To connect the above approximation with uncertainty in a minimal form, let $a_o:=\tilde C(o)>0$ and $b_o:=\Delta I(o)=I_c-I_s(x^\star(o))$, and for $\alpha>0$ define $p_\alpha(o)=\frac{a_o e^{\alpha b_o}}{Z(\alpha)}$ with $Z(\alpha)=\sum_o a_o e^{\alpha b_o}$ and $\Psi_\alpha(r,q)=-\sum_o p_\alpha(o)\log p_\alpha(o)$. Differentiating (using $\frac{d}{d\alpha}\log Z=\mathbb{E}_{p_\alpha}[b]$ and $\frac{d}{d\alpha}\mathbb{E}_{p_\alpha}[f]=\mathrm{Cov}_{p_\alpha}(f,b)$) gives
\begin{equation}
\frac{d}{d\alpha}\Psi_\alpha(r,q)
= -\,\alpha\,\mathrm{Var}_{p_\alpha}(b)\;-\;\mathrm{Cov}_{p_\alpha}(\log a_o,\,b)\ \le\ 0,
\end{equation}
with strict inequality unless $b_o$ is constant. Since $a_o$ varies slowly, the variance term dominates, so increasing the contrast $|\Delta I|$ makes $p_\alpha$ more peaked and lowers $\Psi_\alpha(r,q)$. Averaging over $p(q)p(r\mid q)$ implies $\mathbb{H}(O\!\mid\!R,Q)$ decreases in expectation. (Monotonicity relies on the derivative above; the linear proxy $\exp(\Delta I)\approx 1+\Delta I$ is only for intuition.)

\subsection{Uncertainty Drop and Information}
Without retrieval, the LLM outputs a baseline distribution $p_0$. After adding a retrieved context $r$, the output becomes $p$.
Let $p_0(o)=p(o\mid q)$ and denote $p(o\mid q,r)$ by $p(o)$. Define
\begin{equation}
Z(o)=\log p(o)-\log p_0(o),\quad o\sim p_0.
\end{equation}
By Jensen’s inequality,
\begin{equation}
\mathbb{E}_{p_0}[Z]\le 0
\quad\Rightarrow\quad
U:=-\mathbb{E}_{p_0}[Z]=\mathrm{KL}\!\bigl(p_0\Vert p\bigr)\ge 0,
\end{equation}
where $U$ is a computable proxy that measures how far the augmented output $p$ moves from the baseline $p_0$.
According to instance-level uncertainty $\psi(u)=-u\log u$ for $u\in[0,1]$, and the entropy of a distribution as $\Psi(p)=\sum_o \psi\bigl(p(o)\bigr)$.
The entropy change is
\begin{equation}
\Delta\Psi \;=\; \Psi(p)-\Psi(p_0)
\;=\; \sum_o \bigl[p_0(o)\log p_0(o) - p(o)\log p(o)\bigr],
\end{equation}
which indicates whether augmentation makes the model more confident ($\Delta\Psi<0$).
Consistent with Sec.~2 of the main manuscript, when the information difference $\lvert\Delta I\rvert$ between conflicting and supplementary information is larger, the deviation
$U=\mathrm{KL}(p_0\Vert p)$ tends to increase and we typically observe an entropy drop ($\Delta\Psi<0$).
Thus, a marked decrease in entropy serves as a practical indicator that the  information difference has increased,
even though $U$ (an output-space proxy) is not the same as the latent-space quantity $\lvert\Delta I\rvert$.

\section{Implement Details}

\subsection{Dataset Processing Details}
\label{sec:appendix:dataset}
\subsubsection{ConflictQA}
ConflictQA \cite{xie2023adaptive} is a benchmark for studying knowledge conflicts. This dataset provides LLM’s internal memory and information supporting the external context for each query. To explore the information differences from different ratios in empirical verification of LLM’s preference, it provides two supplementary contexts supporting the LLM’s internal memory ("parametric memory", "parametric memory aligned evidence") and two conflicting contexts supporting the counter answer ("counter memory", "counter memory aligned evidence") for every query.
\subsubsection{DRUID}
DRUID \cite{hagstrom2024reality} contains real-world (query, context) pairs to facilitate studies of context usage and failures. For a multiple-choice task using the DRUID dataset, for each unique "claimant", we select two mutually exclusive "claims" from the dataset as the answer options. Each selected "claim" must have at least one evidence entry marked as either supports or insufficient-supports. The corresponding evidence entries serve as the context for each answer option. The evidence associated with the verified claim is treated as the supplementary context, whereas the evidence for the refuted claim is treated as the conflicting context. If a "claimant" lacks two conflicting claims, we skip that "claimant" and proceed to the next. The ground truth for each question is determined by the "factcheck-verdict": the "claim" with "factcheck-verdict=True" is considered correct. If both claims have "factcheck-verdict=False", we randomly select one "claim" as the ground truth to maintain consistency in the evaluation.
\subsubsection{TruthfulQA}
TruthfulQA \cite{lin2021truthfulqa} evaluates LLM truthfulness across 817 questions, each paired with one best, several correct, and several incorrect answers. Specifically, to implement seamless integration, an Elasticsearch-based retrieval system \cite{elasticsearch} with an embedding model named m3e-base \cite{M3E} and a database named TruthfulQA is constructed. For each query, we retrieve the top five correct or incorrect responses to fill the slot <external context> in the corresponding templates (see Figure \ref{fig:experiments:dataset}). Among these retrieved candidates, the one selected by our LLM  is the supplementary context, while unselected candidates collectively form the conflicting context. Finally, we use the "best answer" key in TruthfulQA as the ground truth for evaluating Swin-VIB's performance in RAG.

\subsubsection{Mixed Context}
To provide the mixed contexts, a supplementary context and a conflicting context are sampled from the dataset and interleave with each other by every four tokens, forming a mixed context.

\subsubsection{Simulated Conflict Generation}
\begin{table}[h]
\renewcommand\arraystretch{0.6}
\centering
\small
\caption{Dataset information}
\label{tab:dataset}
\setlength{\tabcolsep}{0.2mm}{
\begin{tabular}{cccc}
\hline
\multirow{3}{*}{Dataset} & \multirow{2}{*}{Size} & \multicolumn{2}{c}{Task}                                                          \\ \cline{3-4} 
                         &                       & Multiple-choice             & Open-ended QA                                       \\ \hline
ConflictQA    & 2839                  & \checkmark &                           \\ \hline
DRUID         & 1003                  & \checkmark &                            \\ \hline
TruthfulQA    & 817                   &            & \checkmark          \\ \hline
\end{tabular}}
\end{table}
\begin{figure}[t]
  \includegraphics[width=\columnwidth]{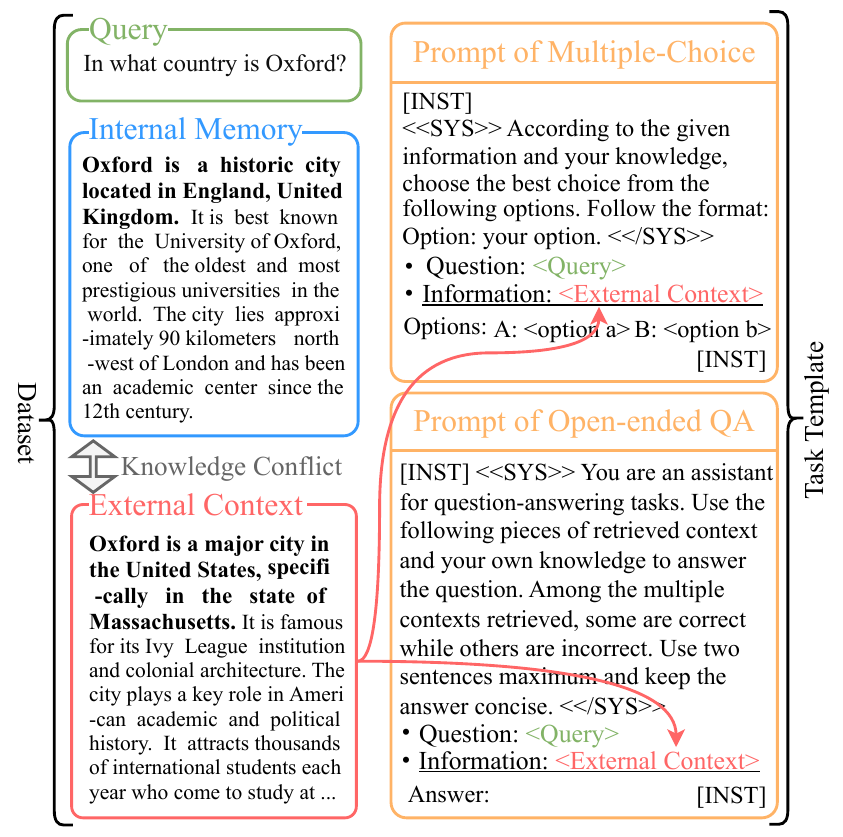}
  \caption{Our framework for simulating knowledge conflicts.}
  \label{fig:experiments:dataset}
\end{figure}
For LLMs not included in the dataset, we uniformly construct conflict scenarios in the following manner: First, each query from the dataset is posed to the LLM in a closed-book setting, so the LLM’s response and accompanying rationale are treated as its bona fide internal memory; Second, we prompt ChatGPT \cite{openai2023gpt} to draft a coherent passage whose factual content contradicts that of its internal memory. The overall procedure is illustrated in Figure \ref{fig:experiments:dataset}.

\subsection{Hardware and Software Environment}
Experiments were run on Ubuntu 20.04 with NVIDIA RTX A6000 GPUs under PyTorch 2.4.1.

\subsection{Model Configuration}
\label{sec:appendix:hyperparameter}
A variational information-bottleneck network that first maps the 3584-dimensional input through a linear layer to 2048 units, followed by batch normalization, ReLU activation, and 50\% dropout. The encoder splits into two parallel linear heads (2,048 → 512), producing the latent mean $\mu$ and log-variance log $\sigma^{2}$. Using the reparameterization trick, the latent vector $\mathbf{Z}$ is sampled and fed into a three-stage decoder: 512 → 256 (with batch normalization, ReLU, and dropout), 256 → 128 (with batch normalization, ReLU, and dropout), and finally 128 → 1. A sigmoid transforms the final logit into a probability. Training minimizes binary-cross-entropy for prediction accuracy plus a KL-divergence term that regularizes $\mu$ and $\sigma^{2}$, weighted by the $\beta$ hyperparameter. The reported results are based on a two-fold cross-validation.
\begin{figure*}[t]
  \centering
  \includegraphics[width=1\linewidth]{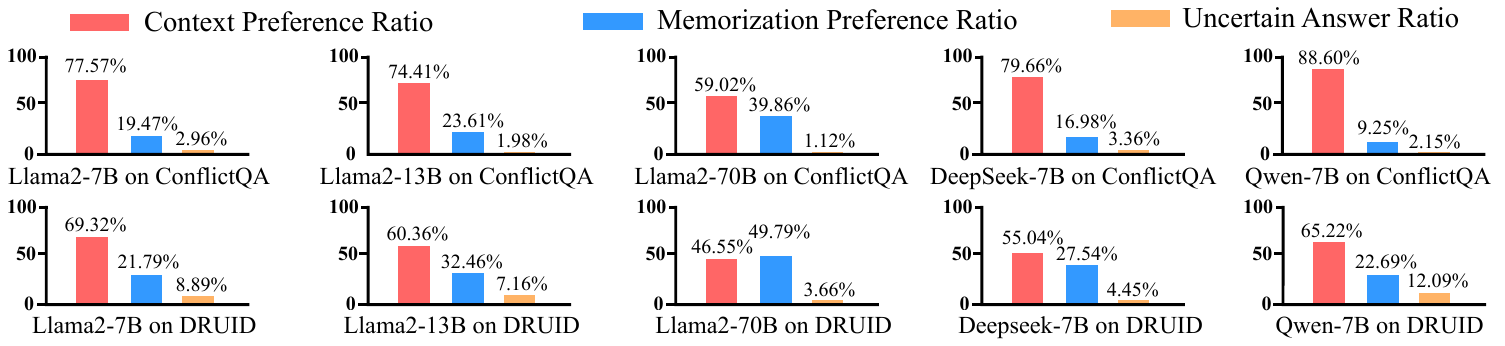}
  \caption{The answer preference distribution of LLM under different datasets between internal memory and external context.}
  \label{fig:detecting:preference}
\end{figure*}

\subsection{Implementation details for each baseline}
We compare Swin-VIB with five baselines on the multiple-choice and plug it into three advanced RAG methods for open-ended QA:
\begin{compactitem}
  \item Closed-book: LLM answers without any external context
  \item In-context: Conflicting context is added to the prompt
  \item CD$^{2}$ \cite{jin2024tug}: Each option undergoes two forward passes, computing log-probabilities on the bare question and with conflicting context; the option with the larger weighted difference is chosen..
  \item CK-PLUG \cite{bi2025parameters}: Evaluation procedures identical to CD$^{2}$.
  \item Rowen-CL \cite{ding2024retrieve}: invokes the Serper Google Search API to retrieve auxiliary Information whenever its consistency score drops below 0.6.
  \item Naive RAG: Based on TruthfulQA, an Elasticsearch-based retriever \cite{elasticsearch} indexed with m3e-base embeddings \cite{M3E} (index name TruthfulQA) is built.
  \item Self-RAG \cite{asai2023self}: Adaptive retrieval threshold = 0.2, ndocs = 5.
  \item Astute RAG \cite{wang2024astute}; Final answer chosen by reliability-weighted voting (default $\tau$ = 0.7) .
\end{compactitem}
For all methods, we use the recommended config for fair comparisons.

\subsection{Instance-level uncertainty}
\label{sec:uncertainty}
Given a query $q$ and a retrieved context $r$, let the language model generate an answer
$o = (o_1,\dots,o_{|o|})$ of length $|o|$ tokens.  
We define the instance-level uncertainty of this generation as the average token-wise negative log-likelihood:
\begin{equation}
\psi(q,r) \;=\;
-\frac{1}{|o|}\sum_{t=1}^{|o|}
\log P\!\bigl(o_t \mid o_{<t},\,q,\,r\bigr),
\label{eq:psi}
\end{equation}
where $o_{<t} = (o_1,\dots,o_{t-1})$ denotes the prefix generated before step~$t$. A more confident LLM places higher probability mass on the realised answer tokens $o_t$, leading to a lower value of $\psi(q,r)$. For a dataset containing $S$ samples $\{(q_i,r_i)\}_{i=1}^S$, we report the
mean uncertainty
\begin{equation}
\text{Mean-}\psi \;=\; \frac{1}{S}\sum_{i=1}^{S}\psi(q_i,r_i),
\end{equation}
which summarises the model’s overall confidence across the corpus.
\section{Supplementary Experiments}

\subsection{Answer-Preference Distribution Across Models and Datasets}
Knowledge conflicts may cause uncertainty in RG because LLMs must choose between internal memory and external context. 
In this paper, we conduct controlled experiments to construct knowledge conflicts.
We evaluated five LLMs on the multiple-choice task, as shown in Figure \ref{fig:detecting:preference}.  Llama-2-7B and Llama-2-70B differ by 22.77\% in CPR and by 28\% in MPR on DRUID. These results reveal substantial cognitive divergence among the dataset's model-specific knowledge representations. Even the same LLM has obvious differences in preferences on different input distributions. For example, Llama-2-70B has different preferences on two datasets, which may be due to the bias of the training corpus. So our controlled experiments reveal that this answer-preference is highly idiosyncratic: it varies not only across models but also across datasets. Consistent with this observation, we deliberately extract attention representations that are specific to each model–dataset pair rather than using a single, global representation. Aligning the representation granularity with the observed preference patterns ensures that our method faithfully captures the factors driving each model’s decisions and avoids spurious, cross-dataset generalization.

\subsection{Training Convergence Analysis}
In this section, we evaluated convergence analysis and layer-wise prediction performance based on the multiple-choice task.
\subsubsection{Convergence Analysis}
\begin{figure}[ht]
\centering
\subfigure[Convergence Analysis on Bottleneck 1 of Qwen3-8B]{
\hspace*{0.08\linewidth}
\includegraphics[width=0.8\linewidth]{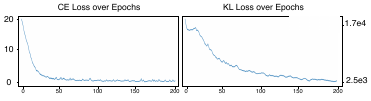}
\label{fig:experiment-bottleneck:a}
}\\
\quad
\subfigure[Layer-wise bottleneck analysis of Llama 2-7B]{
\includegraphics[width=0.9\linewidth]{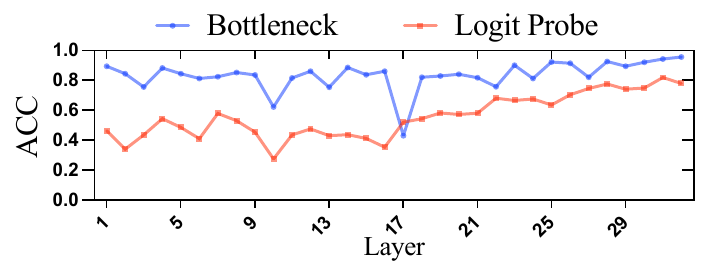}
\label{fig:experiment-bottleneck:c}
}
\caption{Convergence Analysis.}
\label{fig:experiment-bottleneck}
\end{figure}
Figure \ref{fig:experiment-bottleneck:a} depicts the convergence of the bottleneck module trained on Qwen3-8B’s first block on the ConflictQA: the information-bottleneck loss $\mathcal{L}$ plateaus after $\approx$ 200 epochs, evidencing rapid optimization and confirming the loss term’s effectiveness. 
\subsubsection{Layer-wise Prediction Performance}
For each decoder block, we project the last-token hidden state through the tied LM head to obtain logits, then apply softmax, and compare the resulting log-likelihoods of the answer tokens (e.g., “A” vs. “B”). The higher-scoring token yields a depth-indexed multiple-choice prediction. At the same time, we evaluate the bottleneck training on a validation dataset. As shown in Figure \ref{fig:experiment-bottleneck:c}, we observe that the layer-wise information-bottleneck heads consistently outperform the vanilla LLM outputs. The gains occur because the bottleneck isolates high-dimensional discrepancy features that are strongly correlated with epistemic uncertainty. In addition, deeper layers of the LLM handle knowledge conflicts more accurately and with lower variance; however, a few layers (e.g., layers 10 and 17) display pronounced outlier behavior. We advocate weight-averaging the layer-level predictions when forming the final answer to mitigate the latter's adverse influence at test time.

\subsection{Ablation and Sensitivity Studies}
In this section, we evaluated $\beta$-Scaling, sliding-window length study, and Sensitivity to the threshold $\xi$ based on the multiple-choice task.
\subsubsection{$\beta$-Scaling}
\begin{figure}[ht]
\centering
\subfigure[Bottleneck Analysis on ConflictQA]{
\includegraphics[width=5cm]{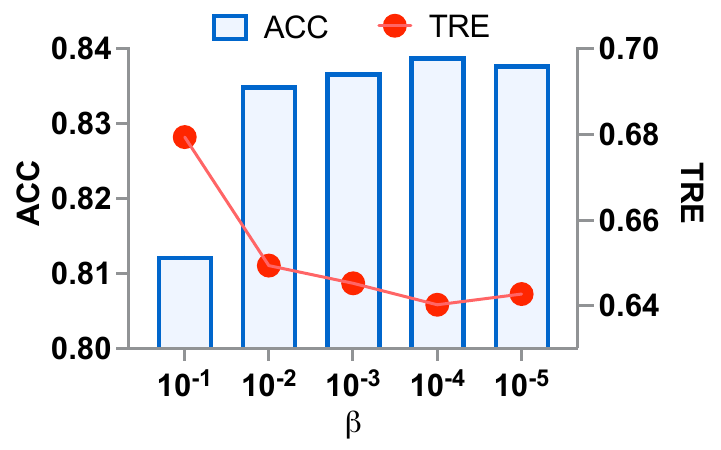}
}\\
\quad
\subfigure[Bottleneck Analysis on DRUID]{
\includegraphics[width=5cm]{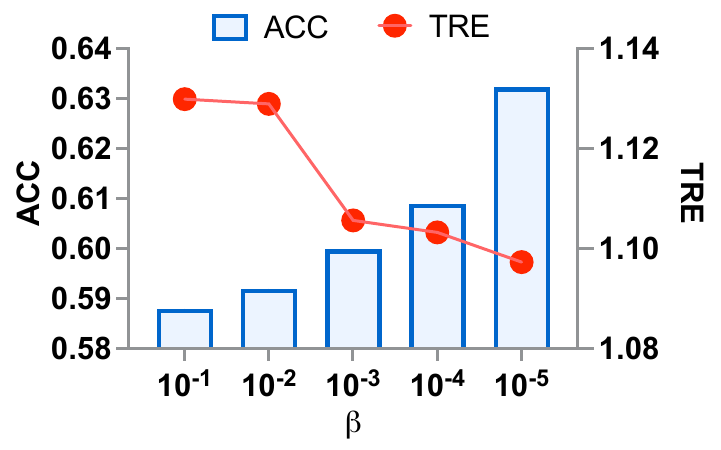}
}
\caption{Parameter analysis.}
\label{fig:experiment-beta}
\end{figure}
\begin{table}[ht]
  \centering
  \caption{Cross-dataset transfer of the threshold $\xi$. $ACC_{\text{self}}$ is the accuracy obtained with each dataset’s own optimal threshold $\xi_{\text{self}}$, while $ACC_{\xi=0.68}$ is the accuracy when the fixed threshold $\xi=0.68$ is applied.}
  \label{tab:xi-transfer}
  \begin{tabular}{lccc}
    \toprule
    \textbf{Dataset} & $\xi_{self}$ & $ACC_{self}$ &  $ACC_{\xi=0.68}$\\
    \midrule
    DRUID           & 0.5        &0.6324       &0.5747\\
    TruthfulQA      & 0.7        &0.6970       &0.6710\\
    \bottomrule
  \end{tabular}
\end{table}
\begin{table*}[ht]
\centering
\caption{Information-type distribution in accepted vs.\ rejected windows.}
\begin{tabular}{lcccc}
\toprule
Group & Conflict & Supplement & Undecidable & Undecidable (\%) \\
\midrule
Accepted by \textsc{Swin-VIB} & 45 & 40 & 15 & 15.0 \\
Rejected / Original retrieval & 12 & 12 & 76 & 76.0 \\
\bottomrule
\end{tabular}
\label{tab:human-info-gap}
\end{table*}
We sweep the bottleneck parameters $\beta$ over five orders of magnitude $(10^{-1}\!\rightarrow\!10^{-5})$ and evaluate ACC together with total TRE on ConflictQA and DRUID in Figure \ref{fig:experiment-beta}. In both corpora, ACC rises while TRE falls as $\beta$ decreases, indicating a clear negative correlation between predictive accuracy and epistemic uncertainty. The trend suggests that tighter bottlenecks i.e., stronger compression of the latent representation force the encoder to discard redundant dimensions and focus on the task-relevant high dimensional discrepancies between conflicting contexts, thereby improving classification while reducing uncertainty.
\subsubsection{Sliding-Window Length Study}
\begin{figure}[ht]
\centering
\includegraphics[width=8cm]{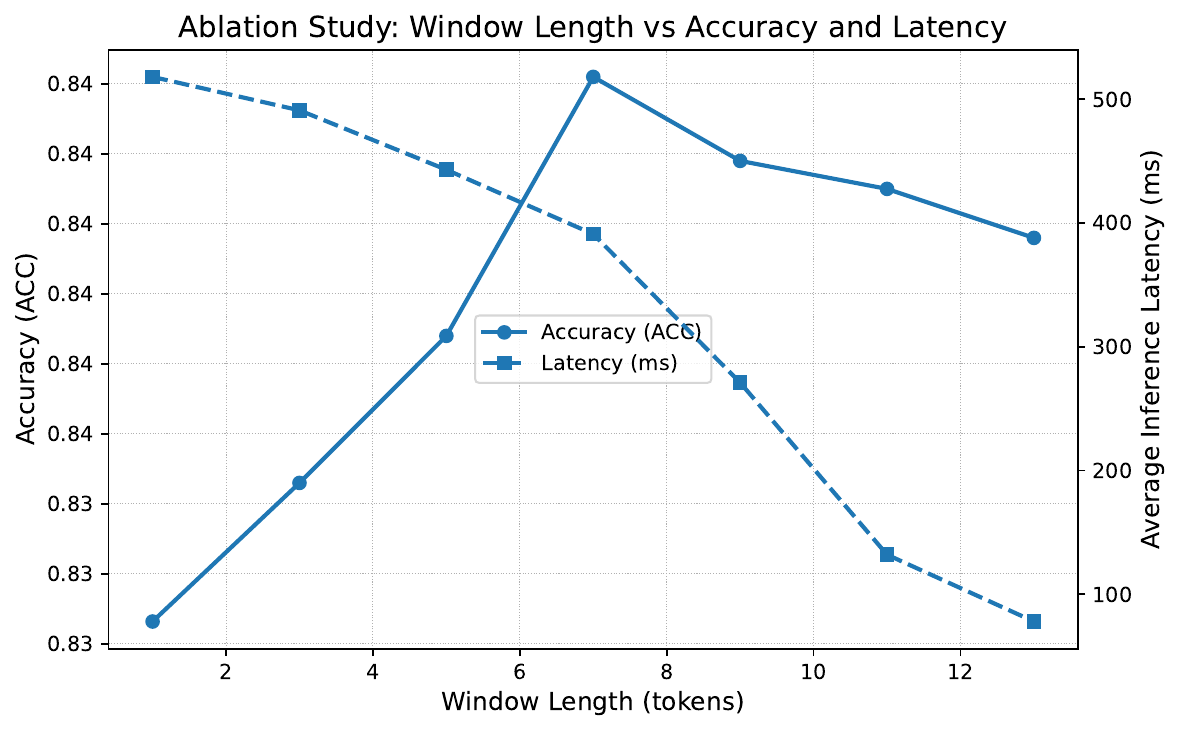}
\caption{Ablation study of sliding-window length on uncertainty prediction accuracy (left axis) and average inference latency (right axis) on Llama2-7B based on ConflictQA.}
\label{window}
\end{figure}
\begin{figure}[ht]
  \centering
  \includegraphics[width=0.8\linewidth]{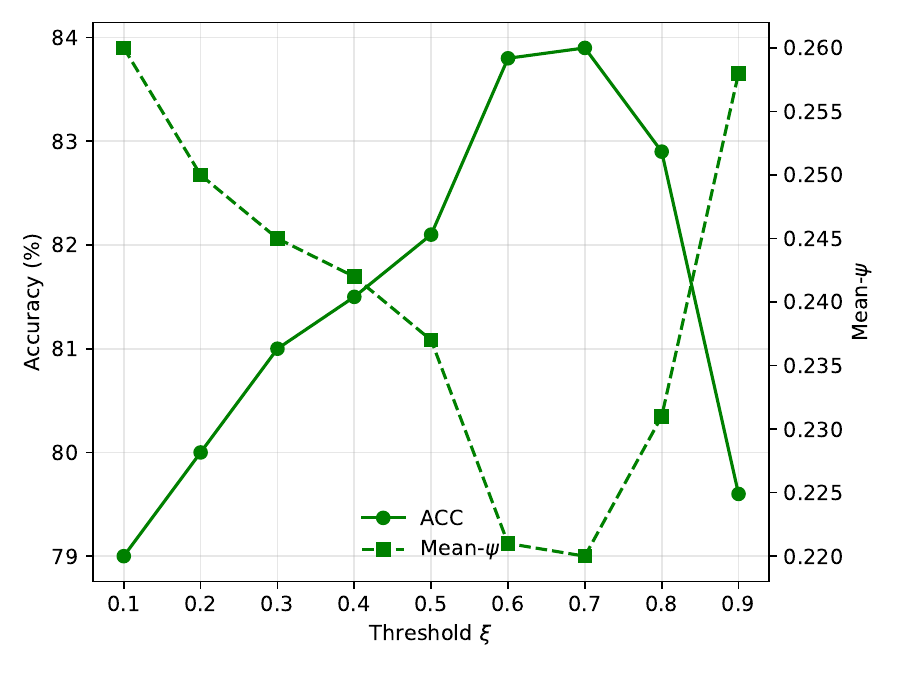}
  \caption{Accuracy (solid) and Mean-$\psi$ (dashed) as a function of the threshold~$\xi$.}
  \label{xi}
\end{figure}

As shown in Fig. \ref{window}, we varied the sliding-window length from 1 to 13 tokens and observed a clear trade-off between prediction quality and computational cost. When the window contains too few tokens (1–3), each inference pass carries insufficient semantic information, resulting in low accuracy in predicting uncertainty. As the window grows, accuracy rises sharply and reaches its maximum at around 7 tokens, where each window strikes an optimal balance between contextual richness and inference complexity. Beyond this point, accuracy declines modestly because a single inference must process overly complex information, which increases the risk of erroneously rejecting relevant content. Meanwhile, inference latency decreases steadily with larger windows, since fewer passes are required over the fixed-length context. Based on these results, we select a window length of 7 tokens to achieve the best compromise between accuracy and efficiency.
\subsubsection{Sensitivity to the threshold $\xi$}
\label{app:xi}
Figure~\ref{xi} plots ACC and Mean-$\psi$ versus~$\xi$;
We sweep the threshold $\xi$ that accepts context windows from $0.10$ to $0.90$ with a step of $0.1$. For each value, we train Swin-VIB once and evaluate on three datasets. Performance is remarkably flat when $\xi\!\in\![0.6,0.8]$: ACC and Mean-$\psi$ varies within $4\%$. Below $0.50$ the model retains too many noisy windows, inflating uncertainty and hurting ACC by up to 2.3\%. Above $0.80$, aggressive pruning occasionally removes all candidate windows, leading to unstable outputs and a sharp performance drop.
We select $\xi^\star\!=\!0.68$ on ConflictQA and directly transfer it to the other two datasets. The gap to each dataset’s own $\xi$ is $6\%$ ACC, confirming that the threshold generalises across domains (Table~\ref{tab:xi-transfer}).

\subsection{Computational Complexity Analysis}
Assume an input context of length $f$ tokens, a Transformer with $N$ layers and hidden width $d$, and let $m \le f$ be the token count after rejecting. 
Introducing Swin-VIB splits inference into three sequential parts. First, the entire context is passed through the N-layer language model to obtain per-layer attention representation; the quadratic self-attention term dominates this step and therefore costs $\Theta(Nf^{2})$, identical to the baseline.
Second, each layer’s attention scores are sliced into $\lceil f/len \rceil$ windows of width $len$ and processed by a bottleneck model of sizes $d\!\times\!k$ and $k\!\times\!1$.
Because all windows on a layer can be dispatched as a single batched matrix multiply, the filtering wall-time no longer scales with the number of windows; instead it is bounded by the constant $c_{\text{bottleneck}} = N(dk + k^{2})$, which is negligible compared with any quadratic term. 
Third, the pruned context of length $m$ is fed back into the model to generate the final answer, incurring $\Theta(Nm^{2})$ operations. Summing the three parts yields a total complexity of $\Theta(Nf^{2}) + c_{\text{bottleneck}} + \Theta(Nm^{2})$.
In the worst case where almost no tokens are removed $(m\!\approx\!f)$, this expression simplifies to $2\,\Theta(Nf^{2}) + c_{\text{bottleneck}}$; the quadratic term dominates, and the constant bottleneck cost can be ignored.

\begin{figure}[ht]
  \centering
  \includegraphics[width=1\linewidth]{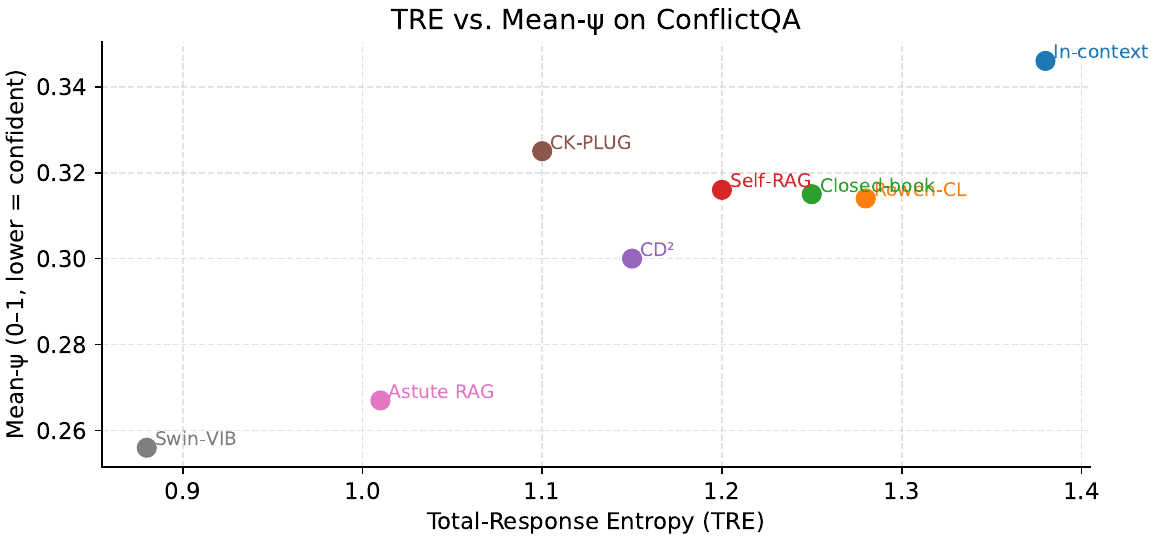}
  \caption{TRE vs.\ Mean-$\psi$ on ConflictQA (each point = a method).}
  \label{fig:tre-vs-psi}
\end{figure}
\begin{figure*}[ht]
    \centering
    \includegraphics[width=1\textwidth]{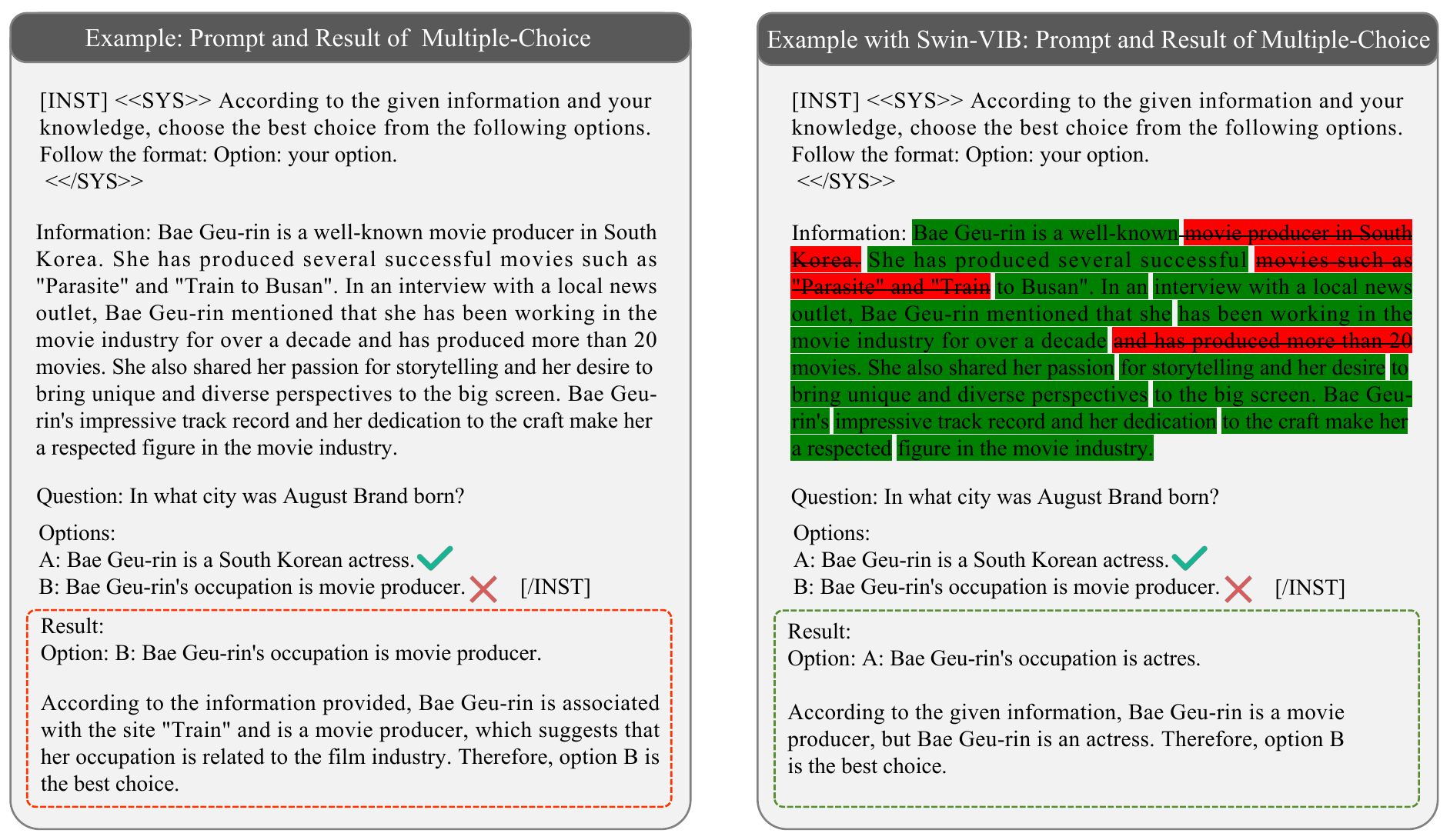}
    \caption{An example of using Swin-VIB on LLama2 for multiple-choice}
    \label{appendix-example-mc}
\end{figure*}
\begin{figure*}[ht]
    \centering
    \includegraphics[width=1\textwidth]{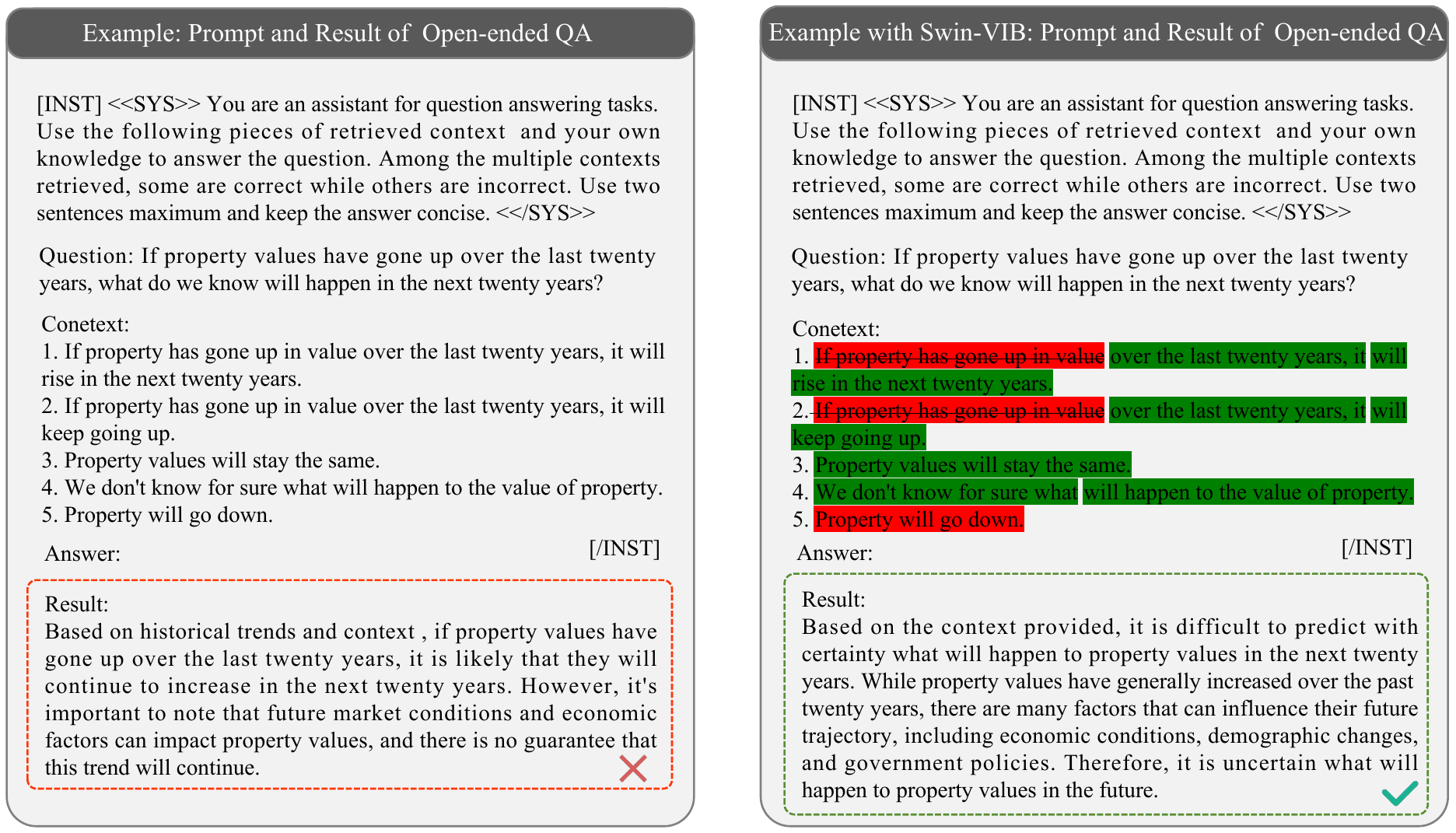}
    \caption{An example of using Swin-VIB on LLama2 for Opened-end QA}
    \label{appendix-example-rag}
\end{figure*}
\subsection{Macro-micro Alignment on \textit{ConflictQA}}
\label{app:corr-conflictqa}
Figure~\ref{fig:tre-vs-psi} plots the macro-level
\emph{Total-Response Entropy} (TRE) against the micro-level
\emph{Mean-$\psi$} for the eight compared methods on
\textit{ConflictQA}.
The Pearson correlation is $r=0.81$,
confirming that the two uncertainty descriptors are tightly coupled:
methods that lower the instance-level uncertainty (\(\downarrow\) Mean-$\psi$) also deliver lower corpus-level entropy (\(\downarrow\) TRE).

\subsection{Real Example}
We randomly drew 100 windows accepted by Swin-VIB and another 100 windows that were rejected from ConflictQA. Each window was labeled as \textsc{Conflict}, \textsc{Supplement}, or \textsc{Undecidable}:
\begin{itemize}
    \item \textsc{Conflict}: information that directly contradicts or refutes the internal memory of LLM.
    \item \textsc{Supplement}: information that supports or extends the internal memory of LLM.
    \item \textsc{Undecidable}: simultaneously contains conflict information and complementary information, or no decisive evidence
\end{itemize} 
Two graduate students with NLP backgrounds independently annotated all 200 windows, blinded to the sampling group. Disagreements (17 \%) were resolved by discussion. Inter-annotator agreement reached Cohen’s $\kappa = 0.78$, indicating substantial consistency.

Table~\ref{tab:human-info-gap} shows that 85 \% of the retained windows contain a single, decisive conflict or supplement cue, while 76 \% of the discarded windows are Undecidable.  
This verifies that Swin-VIB rejects windows of low information difference and keeps those with a larger information difference. Although manual annotation does not yield an exact numeric \(|\Delta I|\), the substantial enrichment of conflict cues in the accepted set provides concrete, human-interpretable evidence that Swin-VIB realizes the intended information difference maximization. 

Figure \ref{appendix-example-mc} (multiple-choice) and Figure \ref{appendix-example-rag} (open-ended RAG) present two representative cases. In both tasks, the vanilla LLM is initially biased toward its internal memory and produces an incorrect answer. After Swin-VIB is introduced, the bottleneck accepts only windows with a large information difference. The LLM’s prediction consequently flips from wrong to correct, providing concrete qualitative evidence predicted by our uncertainty theory.
\section{Administrative Details}
\subsection{Licensing}
\textbf{Dataset Licensing} ConflictQA is released under Apache 2.0; DRUID under the MIT License.\\
\textbf{Model Licensing} Llama 2 is covered by the LLAMA 2 Community License; DeepSeek code is MIT-licensed with model weights under the DeepSeek Model License; Qwen3-8B follows the Tongyi Qianwen License Agreement.
\subsection{Ethical Considerations} 
We do not collect data in this paper. While we generate some auxiliary datasets, we visually inspect the generated examples and do not find any cases of harmful or offensive content. The public datasets used by ConflictQA and DRUID were already vetted by their original creators. All experiments are confined to English-language data and models.

\end{document}